\documentclass[10pt,twocolumn,letterpaper]{article}

\usepackage{cvpr}              

\usepackage{graphicx}
\usepackage{amsmath}
\usepackage{amssymb}
\usepackage{booktabs}
\usepackage{algorithm}
\usepackage{algorithmic}
\usepackage{listings}
\usepackage{ulem}
\usepackage{multirow}
\usepackage{afterpage}
\usepackage{xcolor}
\usepackage{siunitx}

\usepackage[pagebackref,breaklinks,colorlinks]{hyperref}

\definecolor{nblue}{rgb}{0,0.263,0.576}
\newcommand{\authorskip}{\hspace{2.5mm}}
            
\usepackage[capitalize]{cleveref}
\crefname{section}{Sec.}{Secs.}
\Crefname{section}{Section}{Sections}
\Crefname{table}{Table}{Tables}
\crefname{table}{Tab.}{Tabs.}

\def\cvprPaperID{*****} 
\def\confName{CVPR}
\def\confYear{2022}
\begin{document}


\title{Multiplexed Immunofluorescence Brain Image Analysis Using \\ Self-Supervised Dual-Loss Adaptive Masked Autoencoder}

\author{Son T. Ly$^{1}$ \authorskip Bai Lin$^{1}$ \authorskip Hung Q. Vo$^{1}$ \authorskip Dragan Maric$^{2}$ \authorskip Badri Roysam$^{1}$ \authorskip Hien V. Nguyen$^{1}$ \\ [2mm]
$^{1}$University of Houston, Dept. of Electrical and Computer Engineering \\
$^{2}$National Institute of Neurological Disorders and Stroke, Bethesda
}
\maketitle


\begin{abstract}
Reliable large-scale cell detection and segmentation is the fundamental first step to understanding biological processes in the brain. The ability to phenotype cells at scale can accelerate preclinical drug evaluation and system-level brain histology studies. The impressive advances in deep learning offer a practical solution to cell image detection and segmentation. Unfortunately, categorizing cells and delineating their boundaries for training deep networks is an expensive process that requires skilled biologists. This paper presents a novel self-supervised  Dual-Loss Adaptive Masked Autoencoder (DAMA) for learning rich features from multiplexed immunofluorescence brain images. DAMA's objective function minimizes the conditional entropy in pixel-level reconstruction and feature-level regression. Unlike existing self-supervised learning methods based on a random image masking strategy, DAMA employs a novel adaptive mask sampling strategy to maximize mutual information and effectively learn brain cell data. To the best of our knowledge, this is the first effort to develop a self-supervised learning method for multiplexed immunofluorescence brain images. Our extensive experiments demonstrate that DAMA features enable superior cell detection, segmentation, and classification performance without requiring many annotations. Our code is publicly available at \url{https://github.com/hula-ai/DAMA}.
\end{abstract}

\section{Introduction}
Microscopic brain image analysis is critical for medical diagnosis and drug discovery \cite{maric2021whole}. While collecting large amounts of brain imaging data using high-resolution multiplexed microscopes is efficient, annotating these images is time-consuming and labor-intensive. Each brain slice consists of several hundred thousand cells and dozens of cell types. Labeling these images requires highly skilled biology experts. For these reasons, the number of labels for this application is often limited, while the amount of unannotated data is enormous. Self-supervised learning (SSL) offers a practical solution to this situation. 

SSL methods have achieved impressive performance in natural language processing \cite{nlpssl1,nlpssl2,bert}, speech processing \cite{ssl2,speech2}, and computer vision \cite{simclr,moco,dino,data2vec}. SSL aims to learn powerful data representations that are useful for downstream tasks by using pretext tasks created without human supervision. Hence, this approach is ideal for biomedical applications with massive data and limited supervised information. However, there is no best self-supervised method overall \cite{ericsson2021well}; thus, choosing the right SSL learning algorithm is not always straightforward.

\begin{figure*}[t!]
\centering
\subfloat[\centering Overview of DAMA pipeline ]{{\includegraphics[width=0.66\textwidth]{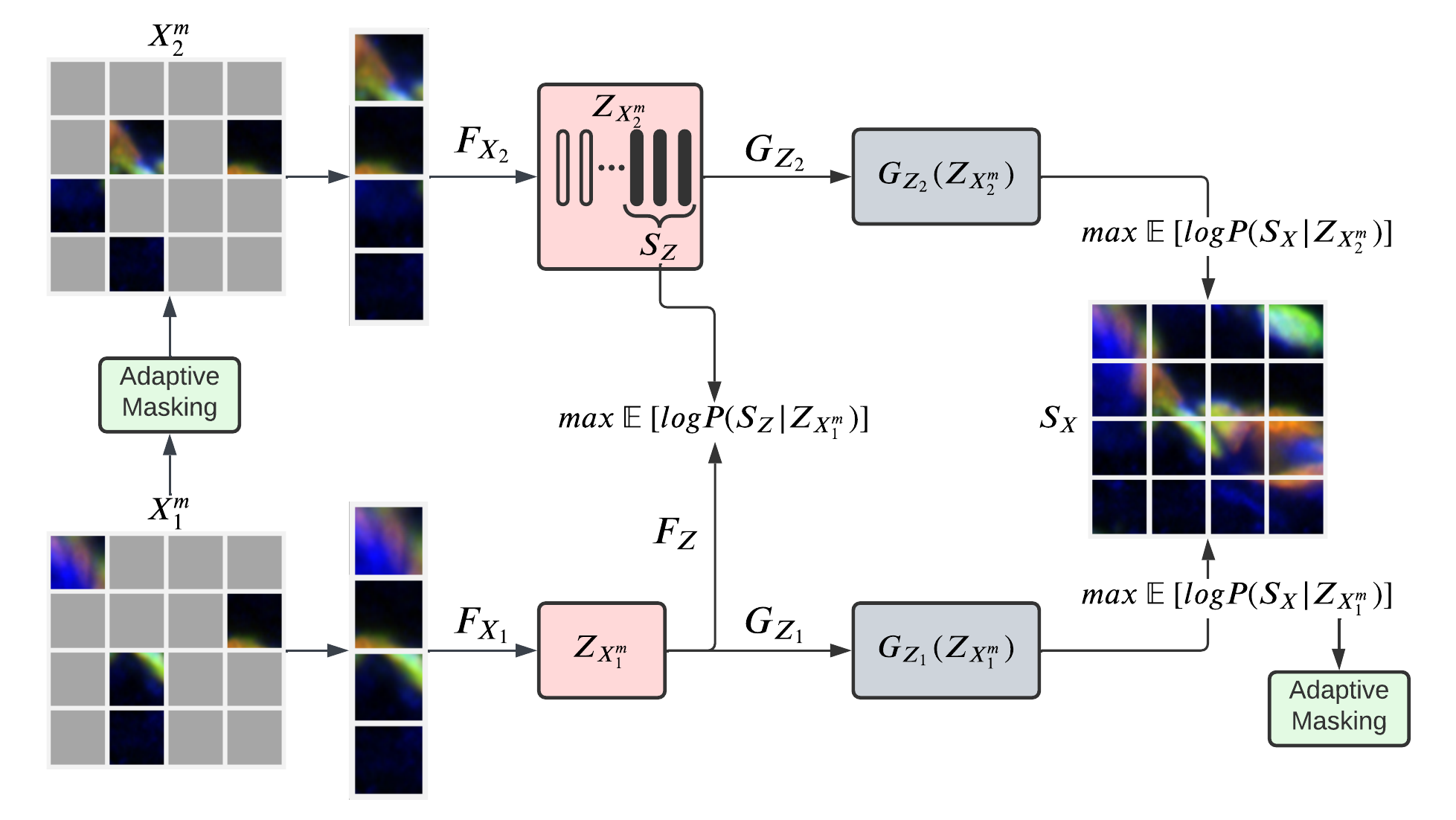}}}\hfill%
\subfloat[\centering Information perspective ]{{\includegraphics[width=0.25\textwidth]{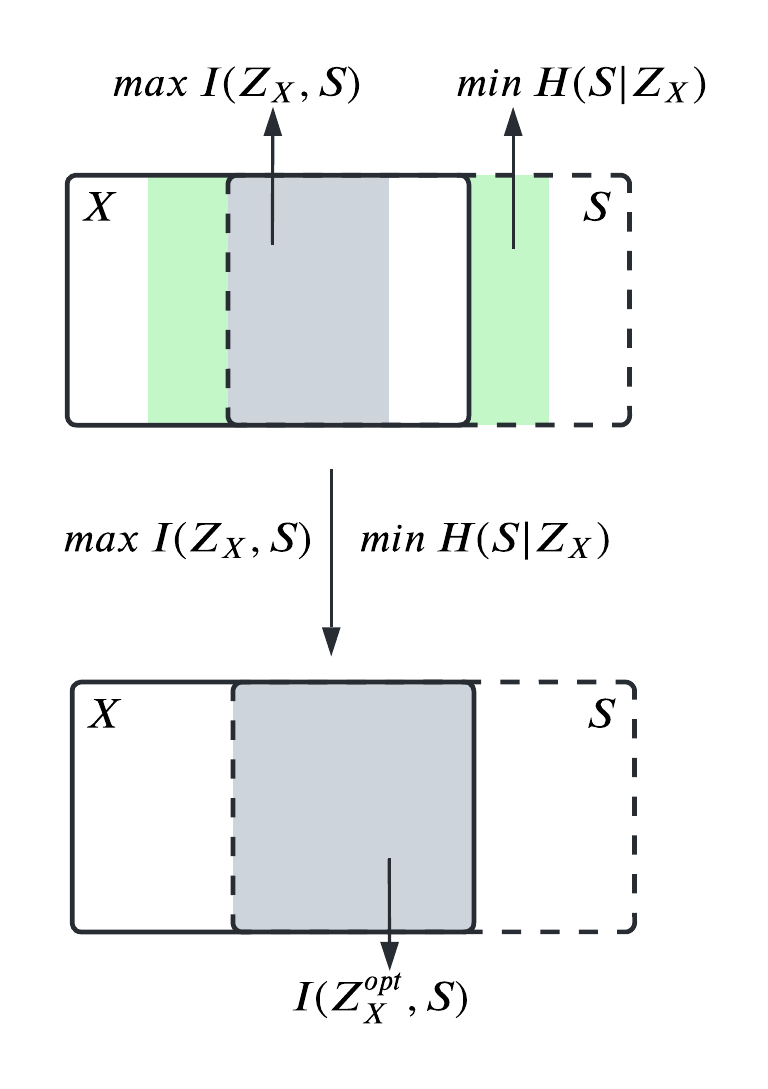}}}%
\caption{(a) Overview of DAMA pipeline. During pre-training, random masked $X_1^m$ input is mapped to the feature space wherein its representations $Z_1^m$ learn to reconstruct the original unmasked input with patch-wise loss (Sec. \ref{subsec:pixel}). The adaptive masking strategy takes the  patch-wise reconstruction loss and the random binary mask of $X_1^m$ and produces the masked input $X_2^m$ (Sec. \ref{subsec:adamask}). It then continues the similar learning process as $X_1^m$. In addition, $Z_1^m$ also try to predict $Z_2^m$ (Sec. \ref{subsec:feature}). The entire procedure encourages maximizing the mutual information between $Z_1^m$ and the self-supervised signal in pixel-level $S_X$ and feature-level $S_Z$ which derive as the log conditional likelihood in the figure. (b) Overview of our pipeline from an information-theoretic perspective. To maximize the mutual information $I(Z_X,S)$, we propose to (1) $min\:H(S|Z_X) = max\:(-H(S|Z_X))=max\:\mathbb{E}_{P_{S,Z_X}}[log\:P(S|Z_X)]$ as pixel-level reconstruction loss \eqref{eq:eq3}, and (2) $max\:I(Z_X,S) = max\:\mathbb{E}_{P_{S,Z_X}}[log\:P(S|Z_X)]$ as feature-level regression loss \eqref{eq:eq4}.}
\label{fig:overview}
\vspace{-5pt}
\end{figure*}

A few recent works have applied SSL to cell data \cite{cell1,cell2,cell3}. For instance, \cite{cell2} reconstructs distorted input to learn representations for quantitative phase image cell segmentation. \cite{cell1} pre-trains one million cancer cell images with a convolutional autoencoder to classify the drug effects. Miscell \cite{cell3} utilizes contrastive learning for mining gene information from single-cell transcriptomes. None of these works studies SSL for multiplexed brain cell analysis. In addition, existing works mainly focus on the applications, while the theoretical analysis for adopting a specific SSL method is often unclear. In contrast, this study aims to bridge the gap between biomedical applications and theoretical motivation. We propose a novel SSL framework that optimizes both pixel \cite{mae,beit,simmim}, and feature-level losses \cite{data2vec,completer,morency} for brain cell image analysis, called DAMA as shown in Fig. \ref{fig:overview}. Our dual loss is motivated by information theory and the observation that the context around cells is useful for analyzing them correctly. The method maximizes the mutual information between masked inputs and self-supervised labels automatically generated by pretext tasks \cite{mae,beit,data2vec}. Specifically, we first mask the original input and then map it to feature space from where the model learns to reconstruct the original unmasked input. Simultaneously, the exact representations also regress to the representations of another masked version of the original input. Joint pixel reconstruction and feature regression increase the consistency of different masked images from the same input, which is the reason behind DAMA's high performance. In addition, from the information-theoretic standpoint, we further propose an adaptive masking strategy to enforce the networks to learn better representations.

The main contributions of our paper are as follows:
\begin{enumerate}
\item We present a novel information-theoretic SSL method for multiplexed brain image analysis. Our theoretical analysis shows that the proposed objective function maximizes the mutual information between the input image and self-supervised labels. 
\item We design the first adaptive mask sampling strategy for self-supervised learning models. We explain how this adaptive sampling helps the model focus on underrepresented features and improve the representation.
\item Extensive experiments on cell detection and classification are provided to validate the effectiveness of DAMA. 
\end{enumerate}

\section{Related Works}\label{sec:relatedwork}
\subsection{Self-Supervised Learning.} Recently, self-supervised learning (SSL) has exhibited a very successful approach in computer vision \cite{simclr,byol,moco,dino,mae,beit}. However, choosing the right SSL learning algorithm is not always straightforward \cite{ericsson2021well}. For example, one of the characteristics of multiplexed biomedical data is that the context also conveys crucial information about the cell. Learning self-supervised signals as multi-view augmented images with contrastive \cite{simclr,byol,moco,morency,completer}, redundancy reduction \cite{twins} or self-distillation \cite{dino} objective will potentially discard this contextual information. As an example, DINO \cite{dino} visualizes the attention maps of Vision Transformers (ViT) \cite{vit} after training and shows that the model mainly focuses on interesting objects while leaving other important information unattended. MAE \cite{mae} and SimMIM \cite{simmim} learn to reconstruct missing image patches from uncorrupted patches. Similarly, Data2Vec \cite{data2vec} regresses the unmasked patches from masked patches at the feature level. However, due to the high random masking ratio, MAE, SimMIM, and Data2Vec would not guarantee to focus on the context information in each iteration. MoCo-v3 \cite{mocov3} learns to increase the mutual information of two augmented views of the same image, e.g., $I(X_1,X_2)$, due to the effect of augmentation transformations, MoCo-v3 would capture the cell body information and abandon the surrounding context distinctive for each cell. Alternatively, based on ViT \cite{vit} framework, we optimize the objective function on both pixel-level reconstruction \cite{mae,simmim,beit} and features-level regression \cite{data2vec} to predict the content of masked regions. By doing so, the algorithm will concentrate on invariant features and the entire image.

\subsection{Masked Image Modeling (MIM).} Recent works built upon Vision Transformer (ViT) \cite{vit} framework, such as BeiT \cite{beit}, MAE \cite{mae}, SimMIM \cite{simmim} have shown potential of learning rich features. These methods propose masking out a random subset of image patches and training deep networks to reconstruct the original image. Our work substantially differs from these methods because it regresses feature representations of multiple ViT blocks in addition to image reconstruction. Data2Vec \cite{data2vec} takes the partially masked images, consisting of masked and unmasked patches, as input and predicts the feature of the original image produced by the second momentum network. In contrast, DAMA takes as the input only visible patches and predicts the features corresponding to visible patches of the second network. All of the current work employs a random masking strategy. On the other hand, DAMA uses an adaptive masking mechanism to learn richer representation and boost performance.

\subsection{Self-Supervised Learning on Biomedical Data} \label{subsec:biomedical}
A few recent works have investigated the applicability of SSL to biomedical data. Motivated by the hierarchical representations of character-, word-, sentence- and paragraph-level in natural language processing, \cite{cell6} introduces a hierarchical pyramid transformer network that can leverage the visual tokens at different sizes the $x_{16}$ cell-, $x_{256}$ patch-, $x_{4096}$ region-level to form slide representations. \cite{cell4} presents an SSL method based on the StyleGAN2 architecture by which the discriminator’s features are exploited for several downstream classification tasks on fluorescent biological images. \cite{cell5} proposes a SimCLR-based method to generate good embeddings of the images for classifying proteomic markers in kidney. While a few papers study multiplexed biomedical data \cite{cell4,cell5}, none of these works sufficiently justify why a specific SSL method reasonably fits their applications. In contrast, this study aims to bridge the gap between biomedical applications and theoretical motivation; and apply it to multiplexed immunofluorescence imaging data.


\section{Self-Supervised Learning from Information Theory Perspective}
\textcolor{nblue}{\textit{Notations.}} For the rest of this paper, we denote the input and self-supervised signal in general as $X$ and $S$, respectively. $S$ can be the augmented image \cite{simclr,moco,dino} or the target of image reconstruction \cite{mae,simmim,beit}. The deterministic mapping function $F$ maps the input $X$ to its representations $Z_X$, i.e. $Z_X = F(X)$, and function $G$ reconstructs the input as $S = G(Z_X)$. We use $I(A,B)$, $H(A)$, and $H(A|B)$ to denote the mutual information, entropy, and conditional entropy of variables $A$ and $B$, respectively.

As shown in Fig. \ref{fig:overview}(b), solid and dotted rectangles represent the information of input $X$ and self-supervised signal $S$, respectively. From the information theory perspective, the mutual information between the representation $Z_X$ and $S$, denoted as $I(Z_X,S)$ (\textcolor{gray}{grey area}), measures the amount of information obtained about one from the knowledge of the other. This mutual information can be expressed as the difference between two entropy terms:
\begin{equation}\label{eq:eq1}
\small
\begin{aligned}
    I(Z_X,S) = H(Z_X) - H(Z_X|S) = H(S) - H(S|Z_X)
\end{aligned}
\end{equation}
In the SSL context, the amount of information of $I(Z_X,S)$ could not be fully inferred, Fig. \ref{fig:overview}(upper b), as $Z_X$ (or $F(X)$) and $S$ are come from random sources, e.g., random augmented views or different random parts of the same image. The goal of SSL is to maximize $I(Z_X,S)$, and consequently, attain better representation about $X$. One can directly maximize $I(Z_X,S)$ like in Completer \cite{completer}. Alternatively, minimizing the conditional entropy $H(S|Z_X)$ (green area) \cite{moco,ssl2} would also encourage $S$ to be fully determined by $X$, indirectly maximizing $I(Z_X,S)$ and minimizing the irrelevant information between $X$ and $S$. Equation \eqref{eq:eq1} can be interpreted as $I(Z_X,S)$ minimize the uncommon information between $X$ and $S$ \cite{morency,twins}. Hence, if $X$ and $S$ are independent, then $I(Z_X,S) = 0$, while if $X$ and $S$ are related, then $I(Z_X,S)$ will be greater than some lower bound. For this reason, $S$ is usually the augmented images \cite{simclr,moco,byol,dino,twins,morency,data2vec} or random masked images \cite{mae,simmim,beit,data2vec}.

Augmentation could boost the performance of self-supervised learning algorithm \cite{simclr,byol,moco}. From the Information Bottleneck principle \cite{ib1}, augmented images could enforce the encoder $F$ to estimate invariant information \cite{mutual}. However, augmentation is data-dependent, and finding the right transformation could sometimes be inconvenient. SimCLR \cite{simclr} conducts a resource-consuming experiment with the combination of only two transformations to find the most favorable combination for ImageNet-1k \cite{imagenet}. Moreover, (strong) augmentations such as cropping, and re-scaling could remove other crucial information, which is important for downstream biomedical tasks. 

Motivated by \cite{morency,completer}, our method aims to minimize the conditional entropy $H(S|Z_X)$.  Fig. \ref{fig:overview}(a) provides an illustration of our method. While \cite{morency} employs forward-inverse predictive learning to boost the performance of contrastive objective, \cite{completer} benefits from dual prediction and contrastive learning for recovering missing views. We, however, take a fundamentally different approach by not targeting to optimize the contrastive function \cite{morency,completer} but focusing on maximizing the mutual information between masked inputs and self-supervised signals at pixel-level reconstruction \cite{mae,beit,simmim} and features-level regression \cite{data2vec}. In addition, while most existing works \cite{mae,beit,simmim,data2vec} utilize a random image masking strategy, our method uses adaptive sampling to more effectively minimize the conditional entropy $H(S|Z_X)$ and learn better representations. To the best of our knowledge, our method is the first to use adaptive image masking for self-supervised learning.
\section{Dual-loss Adaptive Masked Autoencoder}
\label{sec:method}
Motivated by the information theory, this section proposes a Dual-loss Adaptive Masked Autoencoder (DAMA) for SSL. Our method optimizes an objective function associated with pixel- and feature-level information masking. As illustrated in Fig. \ref{fig:overview}(a), it consists of a dual objective function:
\begin{equation}\label{eq:eq6}
\mathcal L_{total} = \mathcal L_{p} + \alpha\:\mathcal L_{f}
\end{equation}
where, $\mathcal L_{p}$ and $\mathcal L_{f}$ are the losses associated with pixel-level reconstruction and feature-level regression, respectively. $\alpha$ is a non-negative constant. From the information theory perspective, our method optimizes $I(Z_X,S)$ and $H(S|Z_X)$. In addition, we present a novel adaptive masking strategy that is better than random masking in terms of performance and theory background. We first provide the context related to the method development and introduce theoretical details later. In our implementation, we fixed $\alpha=1$ for all experiments.

DAMA uses a Vision Transformer (ViT) \cite{vit} as the backbone network. Given an multiplexed input image with seven channels $x\in \mathbb{R}^{H\times W\times7}$, we reshape it into small patches ${(x^P)}^{N}_{i=1}$, where $N = HW/P^2$ patches and $P$ is the resolution of each patch. We masked $m\%$ of the patches and denote them as $\mathcal M = \{1,..., N\}^{m\times N}$. Here, unlike BEiT \cite{beit} and Data2Vec \cite{data2vec} treat the masked patches and unmasked patches as input to ViT, i.e. $x^P = \{x^P_i: i \notin \mathcal M\}_{i=1}^{N} \cup \{e^P_i: i \in \mathcal M\}_{i=1}^{N}$, where $e^P$ is the learnable embedding replacing for masked patches, we feed only the unmasked patches $x^P_U = \{x^P_i: i \notin \mathcal M\}_{i=1}^{N}$ which is similar to MAE \cite{mae}.

\subsection{Pixel-level Reconstruction}\label{subsec:pixel}
Here, we present the theoretical background for the pixel-level loss $\mathcal L_p$ in (\ref{eq:eq6}). In the context of pixel reconstruction, we regard the self-supervised signal $S$ as the reconstruction target, i.e., the original input, denoted as $S_X$ in Fig. \ref{fig:overview}(a).

According to \eqref{eq:eq1}, minimizing the conditional entropy $H(S|Z_X)$ (green area) would also encourage $S$ to be fully determined by $X$, indirectly maximizing $I(Z_X,S)$, and minimization the irrelevant information between $X$ and $S$ \cite{twins,morency}. To do so, the learned representation $Z_X$ is encouraged to reconstruct the self-supervised signal $S$, which leads to maximizing the log conditional likelihood: $-H(S|Z_X) = \mathbb{E}_{P_{S,Z_X}}[log\:P(S|Z_X)]$. However, directly inferring $P(S|Z_X)=\frac{P(Z_X|S)P(S)}{P(Z_X)}$ would be intractable. A common approach to approximate this objective is to define a variational distribution $Q(S|Z_X)$ and maximize the lower bound $\mathbb{E}_{P_{S,Z_X}}[log\:Q(S|Z_X)]$ using variational information maximization technique \cite{varim} as:
\begin{equation}\label{eq:eq2}
\footnotesize
\begin{aligned}
\mathcal I(Z_X,S) &= H(S) - H(S|Z_X) = \mathbb{E}_{P_{S,Z_X}}[log\:P(S|Z_X)] + H(S) \\
                 &= \underbrace{D_{KL}(P(S|Z_X)\:||\:Q(S|Z_X))}_\text{$\geq 0$} \\
                 &\:\:\:\:\:\:+\: \mathbb{E}_{P_{S,Z_X}}[log\:Q(S|Z_X)] + H(S) \\
                 &\geq \mathbb{E}_{P_{S,Z_X}}[log\:Q(S|Z_X)]
\end{aligned}
\end{equation}
We present the $Q(S|Z_X)$ as Gaussian distribution with $\sigma I$ as diagonal matrix, i.e., $\mathcal N(S|G(Z_X),\sigma I)$ \cite{gan,morency,completer}, where $G(\cdot)$ is parameterized as deterministic mapping model which map $Z_X$ to the sample space of $S$. After specifying the mapping functions $G$, the maximizing $\mathbb{E}_{P_{S,Z_X}}[log\:Q(S|Z_{X})]$ objective functions in pixel-level is:
\begin{equation}\label{eq:eq3}
\small
\begin{aligned}
\mathcal L_{p} = 
 \:\:\:min\:\sum_{i=1}^{n}\left[\mathbb{E}_{P_{S,Z_{X_i^m}}}\left \| G(Z_{X_i^m}) - S \right \|_{2}^{2}\right], \:\:\: i=1,2
\end{aligned}
\end{equation}
where $Z_{X_i^m}$ is the representation of masked input $X_i^m$, i.e. $Z_{X_i^m}=F(X_i^m)$, and $i=1,2$ represents for two branches of models as shown in Fig. \ref{fig:overview}. From the above objective function, we can notice that when $\mathcal L_{p}=0$, i.e., $S$ can be fully determined by $Z_{X^m}$, mathematically, $H(S|Z_{X^m}) = 0$. One common situation is that the $G(F(\cdot))$ becomes the identical mapping $I$, and the network will learn nothing. For this reason, $X^m$ is usually the augmented images or random masking images from the same source to avoid the degenerate solution. Our DAMA employs the image masked autoencoder modeling approach similar to \cite{mae,beit,simmim,data2vec} instead of augmenting the input. To leverage the masking operation to contribute more than just generating random masks, we propose a novel \textit{adaptive masking strategy} that can increase the mutual information $I(Z_X,S)$ and learn better representation, whose details will be explained in section~\ref{subsec:adamask}.

\subsection{Feature-level Regression}\label{subsec:feature}
To further encourage maximizing the mutual information $I(Z_X,S)$, DAMA also consist a feature-level regression objective $\mathcal L_f$ in (\ref{eq:eq6}). In the context of feature-level regression, we prefer self-supervised signal $S$ as the feature target produced by $F_Z(X_2^m)$ and indicate as $S_z$ in Fig. \ref{fig:overview}. DAMA predicts feature representations of masked view $X_2^m$ based on masked view $X_1^m$, i.e., $F_Z:Z_{X_1^m}\mapsto Z_{X_2^m}$, where $F_Z$ is the mapping function. This is different from Data2Vec \cite{data2vec} which predicts feature representations of the original uncorrupted input $X$ based on a masked view $X^m$ in a \textit{student-teacher} setting, i.e., $F_Z:Z_{X^m}\mapsto Z_{X^{ori}}$. Furthermore, the masked patches of $X_2^m$ are decided by adaptive sampling strategy.

Similar to the pixel-level loss, maximizing the mutual information $I(Z_X,S)$ leads to maximizing the log conditional likelihood $\mathbb{E}_{P_{S,Z_X}}[log\:P(S|Z_X)]$. We can introduce a variational distribution $Q(S|Z_X)$ and maximize the lower bound $\mathbb{E}_{P_{S,Z_X}}[log\:Q(S|Z_X)]$. Let $Q(S|Z_X)$ as Gaussian distribution with $\sigma I$ as diagonal matrix, i.e., $\mathcal N(S|F_Z(Z_X),\sigma I)$. The objective function is obtained as:
\begin{equation}\label{eq:eq4}
\begin{aligned}
\mathcal L_{f} = min\; \mathbb{E}_{P_{S,Z_{X_1^m}}}\left \| F_Z(Z_{X_1^m}) - S \right \|_{2}^{2}
\end{aligned}
\end{equation}
where $Z_{X_1^m}$ is the representation of masked input $X_1^m$, i.e. $Z_{X_1^m}=F(X_1^m)$. Note that $F$ and $F_Z$ are two different mapping functions; refer to Fig. \ref{fig:overview} for visual illustration. Given that our application of interest is brain image analysis, where context provides critical information for classification and segmentation tasks, we adopt the smooth L1 loss presented in \cite{data2vec}. Hence, the equation \eqref{eq:eq4} becomes:
\begin{equation}\label{eq:eq5}
\begin{aligned}
\mathcal L_{f} =  \begin{cases} 
\frac{1}{2}(F(Z_{X_1^m})-S)^2/\beta, & \left |F(Z_{X_1^m}) - S\right |\leq \beta \\ 
\left |F(Z_{X_1^m}) - S\right| - \frac{1}{2}\beta, & otherwise
\end{cases}
\end{aligned}
\end{equation}
where $\beta$ is the smoothing from L2 to L1 loss term and depends on the difference between $F_Z(Z_{X_1^m})$ and $S$. In addition, the self-supervised signal $S$ is taken from the last \textit{K} blocks of the second branch of the model before normalization to each block and then averaging similarly as in Data2Vec. In our implementation, \textit{K} =  6 and $\beta$ = 2 for all experiments.
\subsection{Adaptive Masking Strategy} \label{subsec:adamask}
Unlike other works \cite{mae,simmim,data2vec,beit}, we propose an adaptive masking strategy to produce masked images $X_2^m$. This strategy helps increase the mutual information $I(Z_X,S)$ and learn better representations. The method is originated from our observation of the theoretical background presented in the pixel-level reconstruction section \ref{subsec:pixel}. The patches with the highest loss indicate the lowest mutual information $I(Z_{X_1^m},S)$. See Algorithm \ref{alg:adamask}.
\begin{algorithm}[t]
\caption{Pytorch-like Adaptive Masking Pseudocode}
\label{alg:adamask}
\definecolor{codeblue}{rgb}{0.25,0.5,0.5}
\definecolor{codekw}{rgb}{0.85, 0.18, 0.50}
\small
\begin{lstlisting}[language=python]
def adaptive_mask(m1, loss, mask_ratio, overlap_ratio):
    # mask_ratio: masking ratio in [0, 1]
    # m1, m2: binary masks {0, 1}; size[N, L]
    # overlap_ratio: overlap ratio between 2 masks
    # loss: patch reconstruction losses; size[N, L]
    # N: batch size
    # L: total number of patches in images
    
    len_keep = int(L * (1 - mask_ratio)) 
    loss_len = int(L - len_keep * 2) 
    overlap_len = int(len_keep * overlap_ratio) 
    
    # get ids of high-loss patches
    # discard losses of unmasked patches in m1
    loss = loss * m1 
    loss_sorted = argsort(loss)
    loss_ids=loss_sorted[:,-(loss_len+overlap_len):]

    # masked patches of m1 become unmasked patches of m2
    # and vice versa, i.e. m1(1) -> m2(0); m1(0) -> m2(1)
    m2 = where(m1 == 1, 0, 1)
    
    # also assign ids of highest loss m1 to m2 as masks
    m2[arange(m2.shape[0])[:,None], loss_ids] = 1
    
    # overlap of unmasked patches of m1 and m2
    m1_ids = argsort(m1)
    m1_ids = m1_ids[:, :overlap_len]
    m2[:, m1_ids] = 0
return m2
\end{lstlisting}
\end{algorithm}
The proposed strategy takes the random binary mask of $X_1^m$ and the patch reconstruction loss in \eqref{eq:eq3} as inputs. It selects the patches with the highest loss, which indicates the lowest mutual information $I(Z_{X_1^m},S)$ as masked patches for $X_2^m$. Regarding the unmasked patches in $X_1^m$, based on the \textit{overlap ratio}, some will become the unmasked patches, and the rest will serve as masked patches in $X_2^m$. The overlap ratio is fixed at 50\% for all experiments. This ensures that the feature-pixel regression would not be too difficult to predict. One can think of the adaptive image masking strategy as a \textit{collaboration} between two students, where the first student estimates the difficulties of reconstructing different patches, and the second student uses that information to select challenging patches to enhance the performance. Note that we develop DAMA upon ViT framework \cite{vit}. Hence, we compute the reconstruction loss patch-wise, and unmasked patches are not considered in computing loss \cite{mae,data2vec,beit,simmim}. 
\section{Experiments}
\label{sec:exps}
In this section, we validate our DAMA on the multiplexed immunofluorescence brain image dataset and compare its performance to supervised learning and state-of-the-art SSL approaches. See Table \ref{tab:sota}, \ref{tab:segment}, \ref{tab:effclsseg}, and Fig. \ref{fig:vizsegsample}, \ref{fig:segprecall}, \ref{fig:dataeff}. 

\subsection{Experimental Settings}
\subsubsection{Multiplex immunohistochemistry staining and imaging} Existing multi-round multiplex immunohistochemistry (IHC) biomarker screening techniques utilize low-plex panels of directly conjugated antibodies with three to four biomarkers per cycle. Commercial automated systems based on these techniques have mainly been optimized for tumor cell biology. We created a staining methodology using up to 10 well-characterized, immunocompatible, and independently confirmed primary antibodies in a single staining cocktail mixture for effective multiplex imaging of brain tissue \cite{maric2021whole}. Target antigens from rat brains were stained using the antibody panels, and it was shown that neither substantial cross-reactivity nor non-specific binding were present. 

Stained sections are then imaged using an Axio Imager.Z2 10-channel scanning fluorescence microscope. Each labeling reaction was sequentially captured in a separate image channel using filtered light through an appropriate fluorescence filter set and the image microscope field (600 × 600\(\:\mathrm{\text\textmu m}\)) with 5\% overlap, individually digitized at 16-bit resolution using the ZEN 2 image acquisition software (Carl Zeiss). For multiplex fluorescence image visualization, a distinct color table was applied to each image to either match its emission spectrum or to set a distinguishing color balance. The pseudo-colored images were then converted into 8-bit BigTIFF files, exported to Adobe Photoshop, and overlaid as individual layers to create multi-colored merged composites. Microscope field images were seamlessly stitched for the computational image analysis described below and then exported as raw uncompressed 16-bit monochromatic BigTIFF image files for additional image processing and optimization. For more information on our multi-round staining and imaging procedure, readers are directed to \cite{maric2021whole}.

\subsubsection{Brain cell dataset} We collect images from 5 major cell types in rat brain tissue sections: neurons, astrocytes, oligodendrocytes, microglia, and endothelial. Seven biomarkers are applied as the feature channels: DAPI, Histones, NeuN, S100, Olig 2, Iba1, and RECA1. DAPI and Histones are utilized to reveal the cells' locations, while other biomarkers are useful for identifying cell types. All models take 7-channel images as the input. 
No cell detection pre-processing was applied to these images; thus, some patches may contain more than one cell, but only the central cell is what we are interested in, and all other cells are generally considered as the background for the cell type classification task. 
Since cell types highly correspond with specific biomarkers, we only use affine transformations, such as rotating, translating, flipping, scaling, and no intensity transformation.

For pretraining and evaluating the SSL methods on the {\textit{classification task}}, our biologists collected and manually annotate 8000 images (1600 cells images for each cell type). We use 4000 images for SSL pretraining, and we keep the remaining 4000 images for finetuning/testing purposes. To validate the performance of each SSL method, we repeat the experiment ten (10) times and average the results. For each experiment, we randomly shuffle images and then split the set of 4000 images into 60\%/40\%, i.e., 2400/1600, for finetuning/testing sets. We perform finetuning of the entire network. Regarding the dataset for SSL experiments, from the remaining 4000 images, we augmented them to 30000 images, called \textit{Set I}, to perform SSL training. 

To exam our method on large-scale settings, we first cropped the brain slide into 1000$\times$1000 images and performed morphological transformations, i.e., erosion. These images were then applied watershed segmentation to identify the cells' location. From cells' locations, we cropped with the size of 100$\times$100 to get cell images, totaling 200,000 cell images regardless of the cell type. From these images, we further split into two sets, called \textit{Set II} and \textit{Set III}, consisting of 30,000 and 170,000 random cell images, respectively. We keep the number of images in \textit{Set II} the same as the augmented images from \textit{Set I} as 30,000 images. This allows us to investigate the robustness of SSL methods against data pre-processing variations. In addition, the \textit{Set III} provides a large-scale setting for SSL training, as the larger dataset, the better the data structure SSL methods could learn.


Regarding the {\textit{segmentation task}}, our biologists manually collected and annotated 181 images of size $512\times512\times7$. We split these images into the size of $128\times128\times7$, totaling 2896 images. We further split each set of images into train/test sets for evaluating the segmentation task with a ratio of 60/40 (1738/1158 images). In this study, the segmentation task requires segmenting the cell body from the background regardless of the cell type.

\begin{table*}[t]
\footnotesize
\centering
\begin{tabular}{lcccccccccccc}
\hline
\multicolumn{1}{c}{Folds}                 & 0     & 1     & 2     & 3     & 4     & 5     & 6     & 7     & 8     & 9     & Avg. Acc. $\uparrow$ & Error $\downarrow$           \\ \hline
ViT. Random   initialized        & 91.75 & 91.19 & 92.75 & 92.69 & 92.56 & 92.31 & 91.44 & 91.06 & 93    & 91.06 & 91.98(+0.00)  &  8.02      \\ 
CellCaps \cite{maric2021whole}        & 95.31 & 94.75 & 95.37 & 96.37 & 94.68 & 95.62 & 95.5 & 95 & 96.12    & 94.56 & 95.32(+3.34)  &  4.68      \\ \hline
                      & \multicolumn{12}{c}{Pretrained on  Set I}               \\ \hline
Data2vec \cite{data2vec} 800 (4h)        & 90.56 & 90.25 & 91.75 & 92.31 & 91.94 & 92.62 & 91    & 91.38 & 92.5  & 90.88 & 91.59(-0.39) &  8.41        \\
MoCo-v3 \cite{mocov3} 500 (6h)            & 90.94 & 91.5  & 92.38 & 92.38 & 92.56 & 92.12 & 91.25 & 90.94 & 92.69 & 90.75 & 91.75(-0.23) &  8.25      \\
MAE \cite{mae}  800 (4h)             & 94.69 & 93.81 & 95.19 & 95.25 & 95    & 93.56 & 94.62 & 93.88 & 95.44 & 94 & 94.54(+2.56)  & 5.46 \\
DAMA-rand 500 (3h) (ours)           & 94.69 & 94.19 & 94.81 & 95.81 & 94.50 & 94.00 & 94.88 & 94.69 & 95.25 & 94.81  & \underline{94.76(+2.78)} & \underline{5.24} \\
DAMA 500 (5h) (ours)           & 95.5 & 94.5 & 95.69 & 96.25 & 95.56 & 95.44 & 95.62 & 94.94 & 95.69 & 95.25 & \textbf{95.47(+3.49)} & \textbf{4.53} \\ \hline                      
                      & \multicolumn{12}{c}{Pretrained on Set II}                                                         \\ \hline
Data2vec \cite{data2vec} 800 (4h)    & 93.69 & 93    & 93.31 & 94.06 & 93.56 & 94.19 & 93.38 & 93.31 & 93.81 & 93.5  & 93.58(+1.60)  &   6.42    \\
Data2vec \cite{data2vec} 1600   (8h) & 91.5  & 90.69 & 92.81 & 92.38 & 92.62 & 92.62 & 91.56 & 91.94 & 92.19 & 91.5  & 91.98(+0.00)  &   8.02    \\
MoCo-v3 \cite{mocov3} 500 (6h)        & 95    & 94.12 & 95.81 & 96    & 95.75 & 95.19 & 95.12 & 94.44 & 95.5  & 95.19 & \underline{95.21(+3.23)} &  \underline{4.79} \\
MoCo-v3 \cite{mocov3} 1000 (12h)      & 94.44 & 93.62 & 94.38 & 95.19 & 94.69 & 95.25 & 94.12 & 94.38 & 95.25 & 94.31 & 94.56(+2.58) &    5.44     \\
MAE \cite{mae} 800 (4h)         & 94.81 & 94.44 & 94.56 & 94.81 & 94    & 94    & 94.38 & 94    & 94.88 & 93.69 & 94.35(+2.37)  &   5.65    \\
MAE \cite{mae} 1600 (8h)        & 94.38 & 94.19 & 95.12 & 95.19 & 94.44 & 93.94 & 94.12 & 94.19 & 95.44 & 93.94 & 94.49(+2.51)  &   5.51    \\
DAMA 500 (5h) (ours) & 95.62	&94.44&	96.06&	96.56&	95.62&	95.56&	95.88&	95.25&	96&	94.88&	\textbf{95.59(+3.61)} & \textbf{4.41}\\
DAMA 1000 (10h) (ours)     & 94.5  & 94.06 & 95.75 & 95.44 & 94.69 & 94.44 & 94.44 & 94.38 & 95.25 & 94.25 & 94.72(+2.74)  &   5.28     \\ 
\hline
                      & \multicolumn{12}{c}{Pretrained on Set III}                                                        \\ \hline
Data2vec \cite{data2vec} 800 (29h)        & 92.25 & 91.06 & 92.5  & 92.69 & 91.94 & 91.56 & 92.88 & 91.5  & 92.25 & 91.31 & 91.99(+0.01)  &  8.01     \\
MoCo-v3 \cite{mocov3} 500 (48h)            & 95.38 & 94.56 & 95.94 & 95.94 & 95.81 & 95.38 & 95.62 & 95.25 & 95.81 & 95.56 & \underline{95.52(+3.54)} &  \underline{4.48} \\
MAE \cite{mae} 800 (34h)             & 94.81 & 93.5  & 95    & 94.38 & 94.88 & 93.69 & 93.94 & 93.81 & 94.88 & 93.69 & 94.25(+2.27)  &  5.75     \\
DAMA 500 (35h) (ours) &95.38&	94.56&	95.81&	96.38&	95.5&	95.38&	95.88&	95.25&	96&	95.56&	\textbf{95.57(+3.59)}& \textbf{4.43} \\ \hline
\end{tabular}
\caption{Comparisons of finetuning results of DAMA and state-of-the-art SSL methods, ViT. randomly initialized, and CellCaps \cite{maric2021whole} on three data settings in accuracy and error rate. Our DAMA reports stable results over dataset settings compared with other state-of-the-art. Training epochs and training times are listed along with the methods. \textbf{Bold} and \underline{underline} are the {best} and second-best scores in each column, respectively.}
\label{tab:sota}
\end{table*}

\subsection{Implementation Details}
We implemented DAMA using Pytorch. Unless stated otherwise, we trained on ViT-Base using Adam optimizer \cite{adam} with a base learning rate of 0.00015, batch size of 512, image size 128$\times$128$\times$7, ViT-Base patch size 16. Regarding state-of-the-art implementations, we take the officially released code \cite{mocov3,mae,deit} and conduct pre-training with our biomedical data, except for Data2Vec \cite{data2vec}.
We report results of our DAMA and MAE \cite{mae} with masking ratios 80\% and 60\% for Data2Vec \cite{data2vec}. Except for SimMIM-Swin with Swin \cite{swin}, all the methods are used ViT \cite{vit} as backbone architecture. All experiments were done on 4 GPUs of V100 32GB. Pre-training or finetuning experiments of different methods on the same dataset have the same random seed.

\subsection{Comparisons with State-of-the-arts}
\subsubsection{Brain cell dataset} In Table \ref{tab:sota} and \ref{tab:segment}, we compare the finetuning classification and segmentation performances of the pretrained self-supervised models. Our DAMA outperforms other methods in both tasks.

\afterpage{\begin{table*}[t]
\centering
\footnotesize
\begin{tabular}{lcccccc}
\hline
\multicolumn{1}{c}{Methods} & Box mAP & Box mAP@50 & Box mAP@75 & Mask mAP & Mask mAP@50 & Mask mAP@75 \\
\hline
ViT random initialized & 63.4 & 90.8 & 73.9 & 66.7 & 90.9 & 76.1 \\
Swin random initialized & 63.2 & 90.6 & 73.7 & 66.3 & 90.5 & 76 \\
MAE \cite{mae} 800 & 63.8 & 90 & 73.3 & 66.3 & 90.1 & 76.3 \\
MAE \cite{mae} 1600 & 63.7 & 90.8 & 74.8 & 67.1 & \textbf{91.4} & \underline{76.9} \\
MoCo-v3 \cite{mocov3} 500 & 63.1 & 90.2 & 73 & 66.1 & 90.8 & 75.2 \\
MoCo-v3 \cite{mocov3} 1000 & 63.2 & 90.5 & 73.2 & 66.5 & 91 & 75.9 \\
SimMIM-ViT \cite{simmim} 800 & 63.6 & 91.1 & 74.1 & 66.9 & 91.1 & 76.1 \\
SimMIM-Swin \cite{simmim} 800 & \underline{64.2} & \underline{91.3} & \underline{75.1} & 67 & 91.2 & \textbf{77} \\
DAMA 500 (ours) & 64.1 & 91.1 & 74.2 & \underline{67.2} & 91.1 & \textbf{77} \\
DAMA 1000 (ours) & \textbf{64.6} & \textbf{91.4} & \textbf{75.3} & \textbf{67.3} & \underline{91.3} & \textbf{77} \\
\hline
\end{tabular}
\caption{Comparisons of segmentation accuracy with state-of-the-arts on 7-channel brain cell dataset. \textbf{Bold} and \underline{underline} are the {best} and {second-best} scores in each column, respectively.}
\label{tab:segment}
\end{table*}}

\textcolor{nblue}{\textit{Cell classification.}} 
Regarding the SSL performances on \textit{Set I} in Table \ref{tab:sota}, one can notice that DAMA is more accurate than image encoding methods, such as MAE \cite{mae} and Data2Vec \cite{data2vec}, even with random masking setting, denoted as \textit{DAMA-rand}. This is expected since DAMA combines the strengths of both methods. MAE and Data2Vec were pretrained for 800 epochs while DAMA used 500 epochs. MAE reports high optimal masking ratios, ranging from 40\%-90\% on ImageNet-1K \cite{imagenet}. For biomedical datasets, we only pretrained MAE with an 80\% masking ratio for all datasets. Since Data2Vec has a similar approach to our feature-level regression, we implemented Data2Vec with a random patch masking ratio of 60\%. Data2Vec does not produce good results since it only learns to regress from low-dimension features.
Compared to the contrastive learning method, MoCo-v3 \cite{mocov3}, our method reconstructs pixels and regresses features instead of optimizing the model through contrastive learning. Our DAMA significantly outperforms MoCo-v3 (95.47\% versus 91.75\%). Moreover, initializing a classifier with MoCo-v3 representation does not improve the accuracy ($91.75\%$) over a randomly initialized classifier ($91.98\%$). This suggests MoCo-v3 introduces biases on small datasets. Despite using a second network to learn adaptive masking, DAMA's pre-training time (5 hours) is still reasonable compared to MAE's (4 hours). Data2Vec (4 hours) only leverages features for learning, leading to less training time. MoCo-v3's training times (6 hours) are slower than MAE, Data2Vec, and our DAMA adaptive masking method.

Regarding the performances on \textit{Set II and III}, Table \ref{tab:sota} presents the comparisons of DAMA and other methods. Our DAMA surpasses other SSL methods with considerable margins on both settings. It suggests that DAMA is stable for both manually annotated small-scale datasets and large-scale datasets. While Data2Vec \cite{data2vec} does not produce good results since it only learns to regress from low dimension features, MAE \cite{mae} has better results and is comparable to MoCo-v3 \cite{mocov3} and our DAMA. MoCo-v3 has the second-best performance after our DAMA. It shows that MoCo-v3 performs poorly for small pretraining datasets but the performance improves substantially on large datasets. This is expected since contrastive learning learns to increase the mutual information of two augmented views of the same image, e.g., $I(X_1,X_2)$, capturing better cell body information that is invariant across the dataset. However, this is also a downside since MoCo-v3 would abandon other critical information useful for segmentation tasks. In addition, in the cases of large image size or/and the number of channels, contrastive learning methods need a large training batch size and, thus, demand high computational resource \cite{ssl2,byol,simclr,moco,mocov3}. It is worth mentioning that MoCo-v3 took a significantly longer time to train (48 hours) than MAE \cite{mae} or our DAMA (35 hours).

\begin{figure*}[]
\centering
\includegraphics[width=0.9\textwidth]{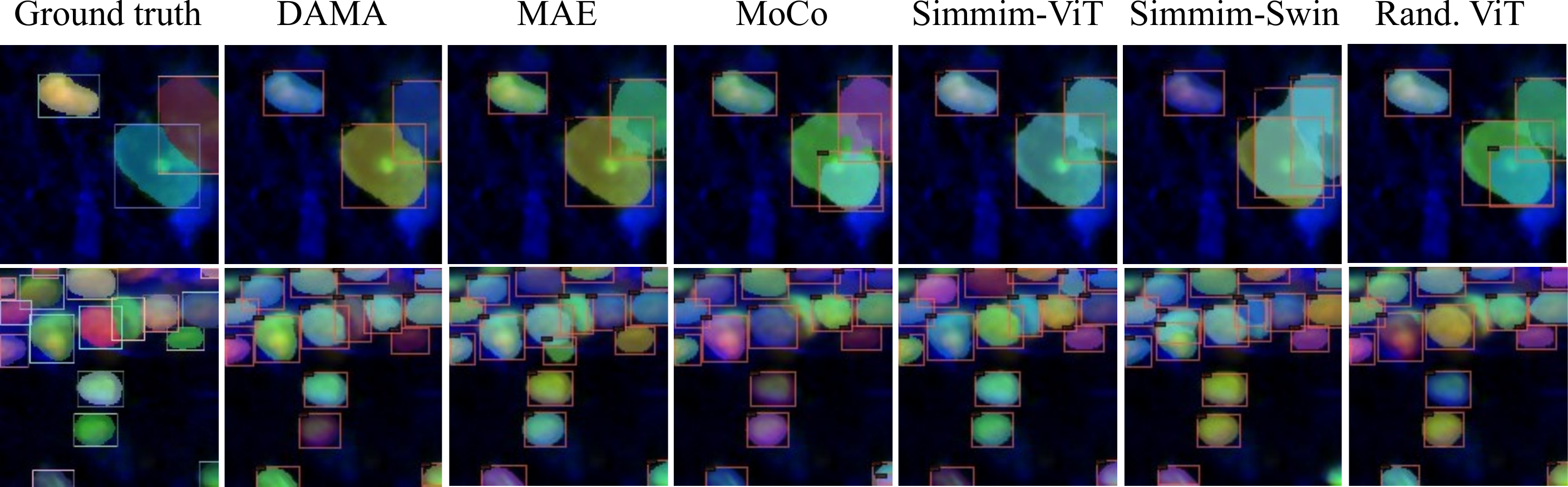}
\caption{Visualization of segmentation results of 7-channel on the validation set at IoU = 0.75. For ease of visualization, images are converted to RGB 3-channels.}
\label{fig:vizsegsample}
\vspace{-5pt}
\end{figure*}

We also compare DAMA to state-of-the-art fully supervised learning methods based on ViT and CellCaps \cite{maric2021whole} -- an equivariant convolutional capsule network designed specifically for learning with a small number of labeled brain cell images. DAMA reduces the error of ViT from 8.02\% to 4.68\%, equivalent to a whopping 41.6\% of error reduction. DAMA performs slightly better than CellCaps, which is also impressive as CellCaps was designed to account for geometric equivariance and to work well with brain cell datasets. We note that training large CellCaps architecture can be unstable due to the iterative dynamic voting mechanism \cite{sabour2017dynamic}, making it challenging to scale to large datasets. Moreover, extending CellCaps to detection/segmentation is not straightforward. In contrast, DAMA can easily adapt to both classification and segmentation tasks as shown in the subsequent sections. Therefore, we believe DAMA is a more scalable, unified, and easy-to-use approach for cell image analysis.

\begin{figure*}[t]
\centering
\subfloat[\centering DAMA (ours) ]{{\includegraphics[width=0.2\textwidth]{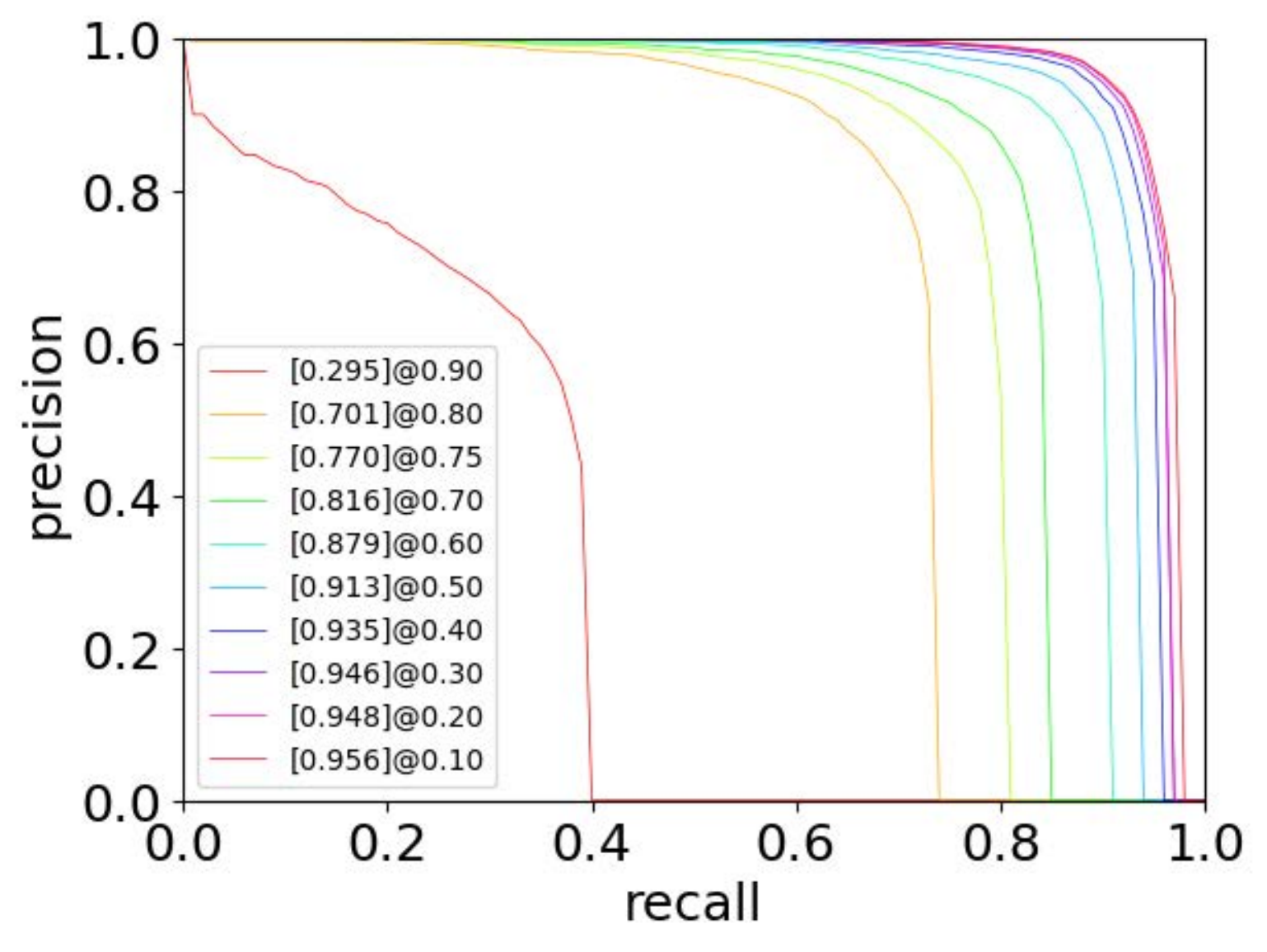}}}\hfill
\subfloat[\centering MAE ]{{\includegraphics[width=0.2\textwidth]{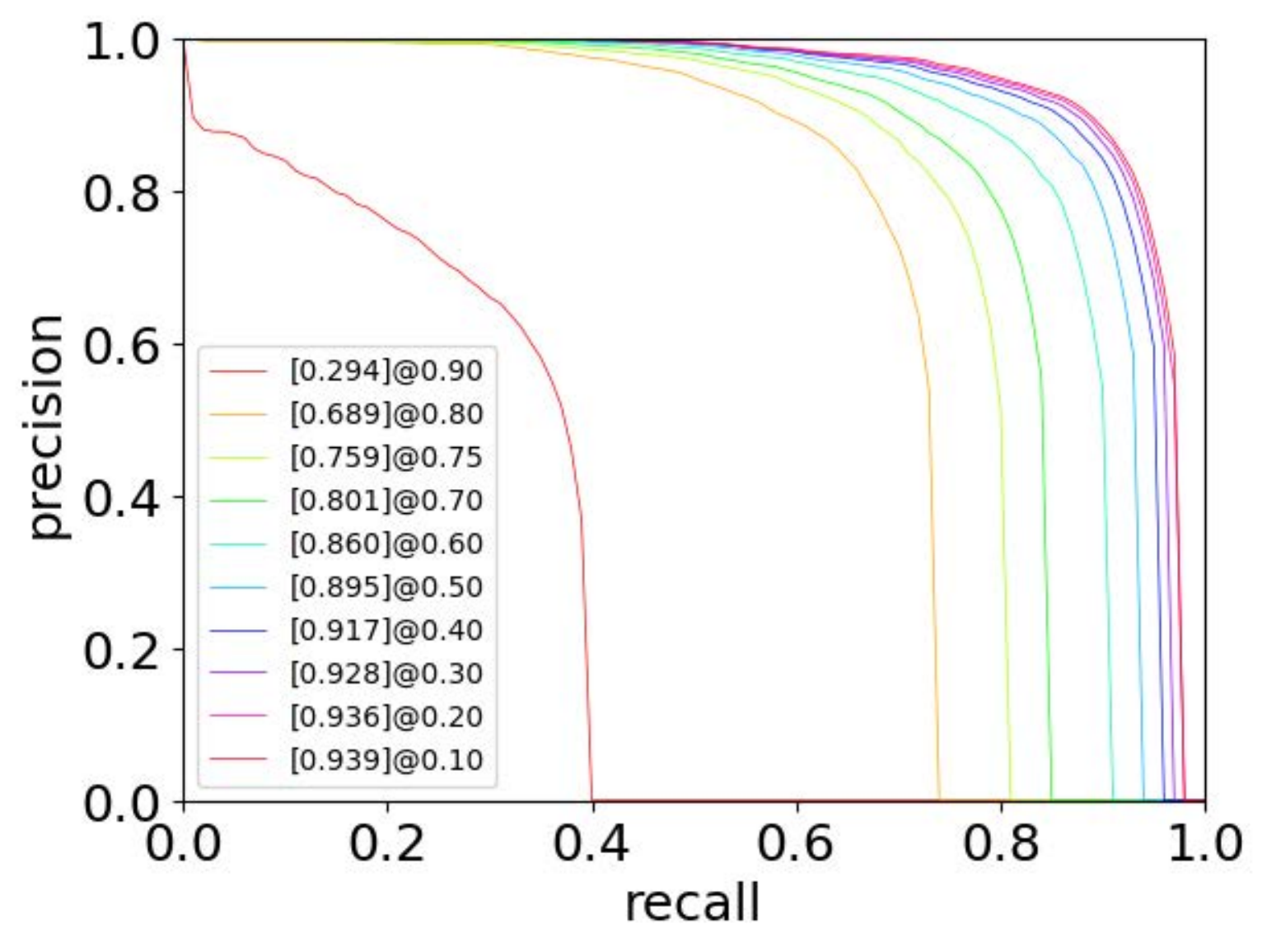}}}\hfill
\subfloat[\centering MoCo ]{{\includegraphics[width=0.2\textwidth]{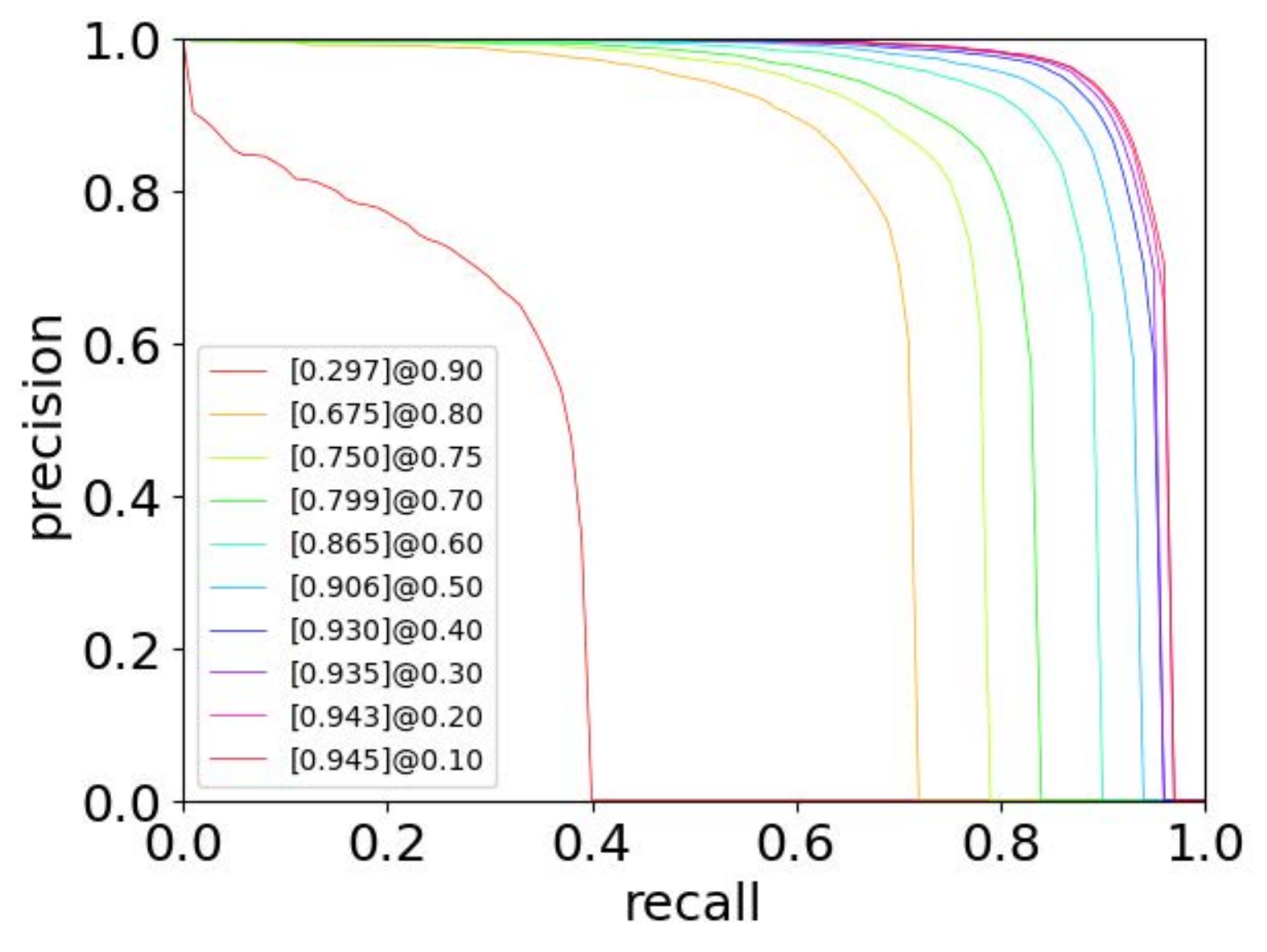}}}\hfill
\subfloat[\centering SimMIM-ViT ]{{\includegraphics[width=0.2\textwidth]{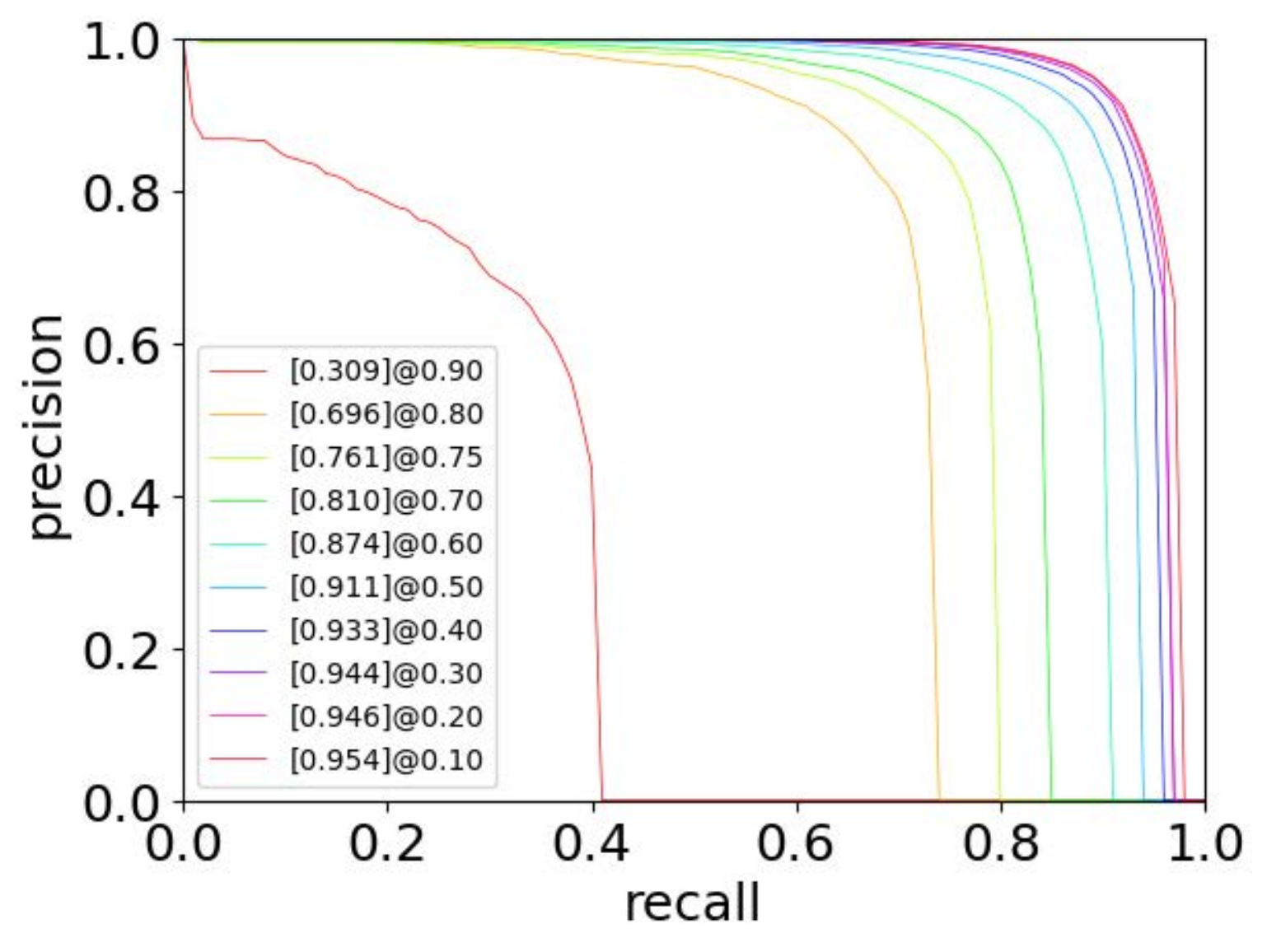}}}\hfill
\subfloat[\centering SimMIM-Swin ]{{\includegraphics[width=0.2\textwidth]{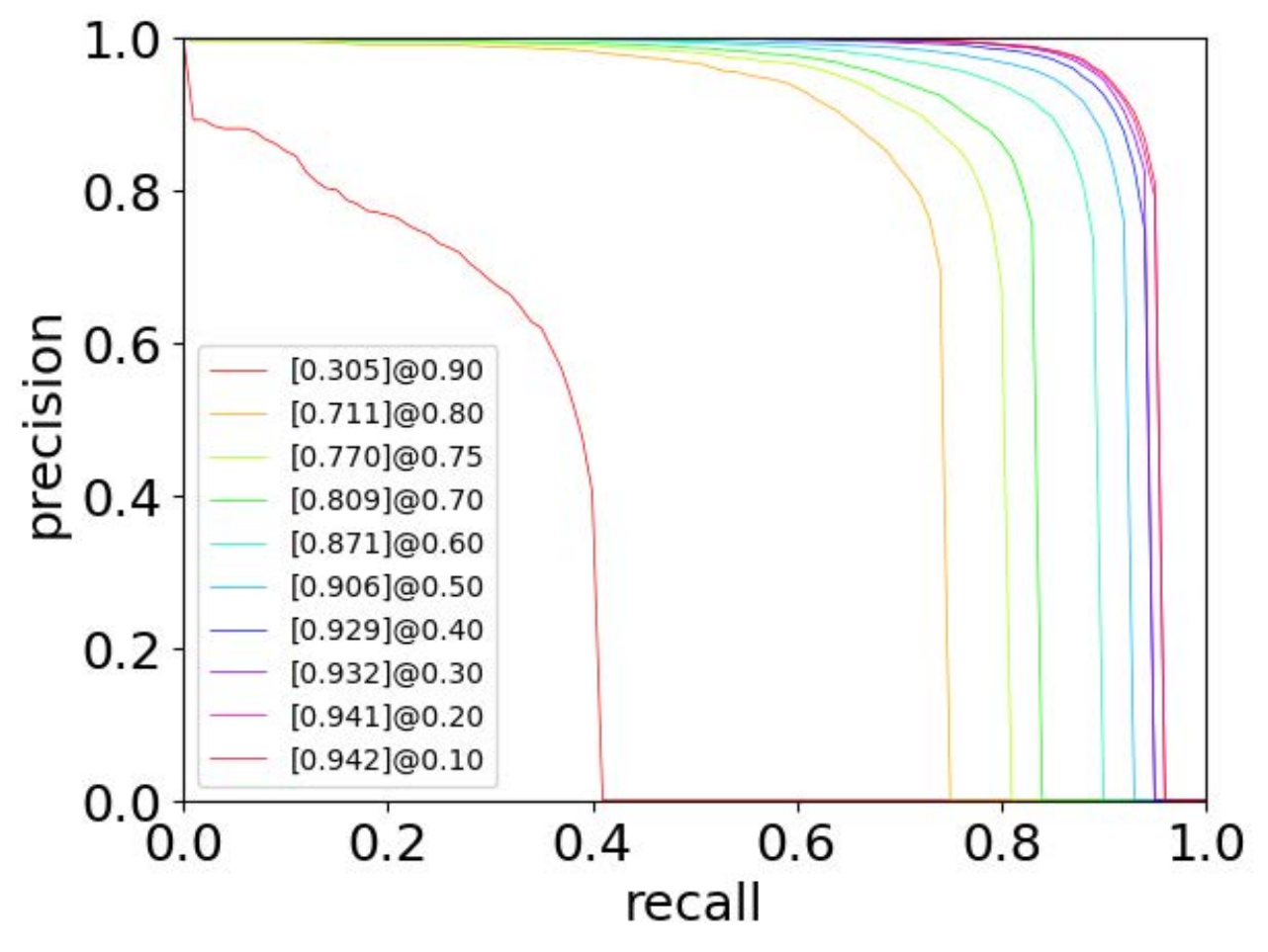}}}\\
\subfloat[\centering DAMA (ours)]{{\includegraphics[width=0.2\textwidth]{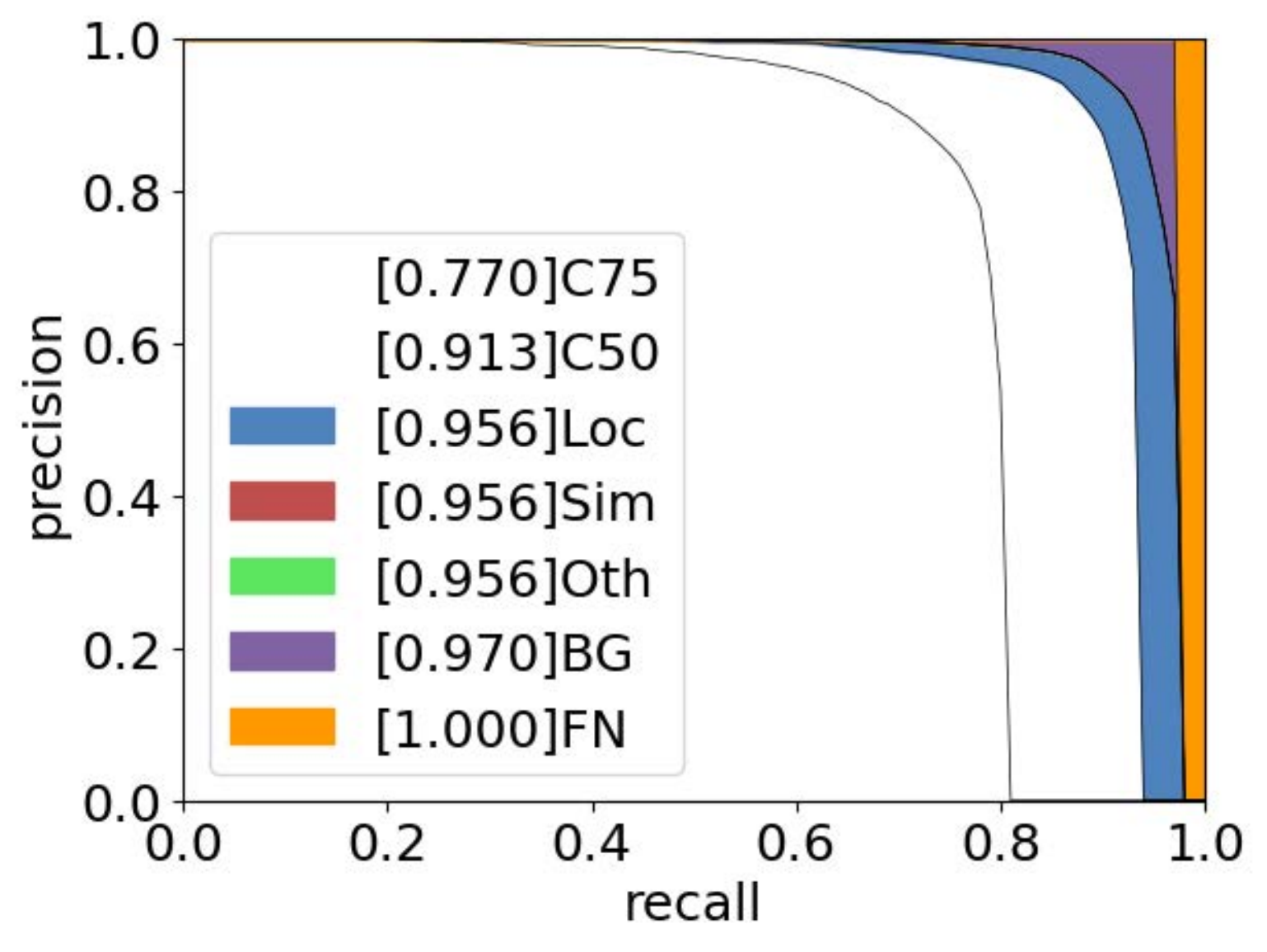}}}\hfill
\subfloat[\centering MAE ]{{\includegraphics[width=0.2\textwidth]{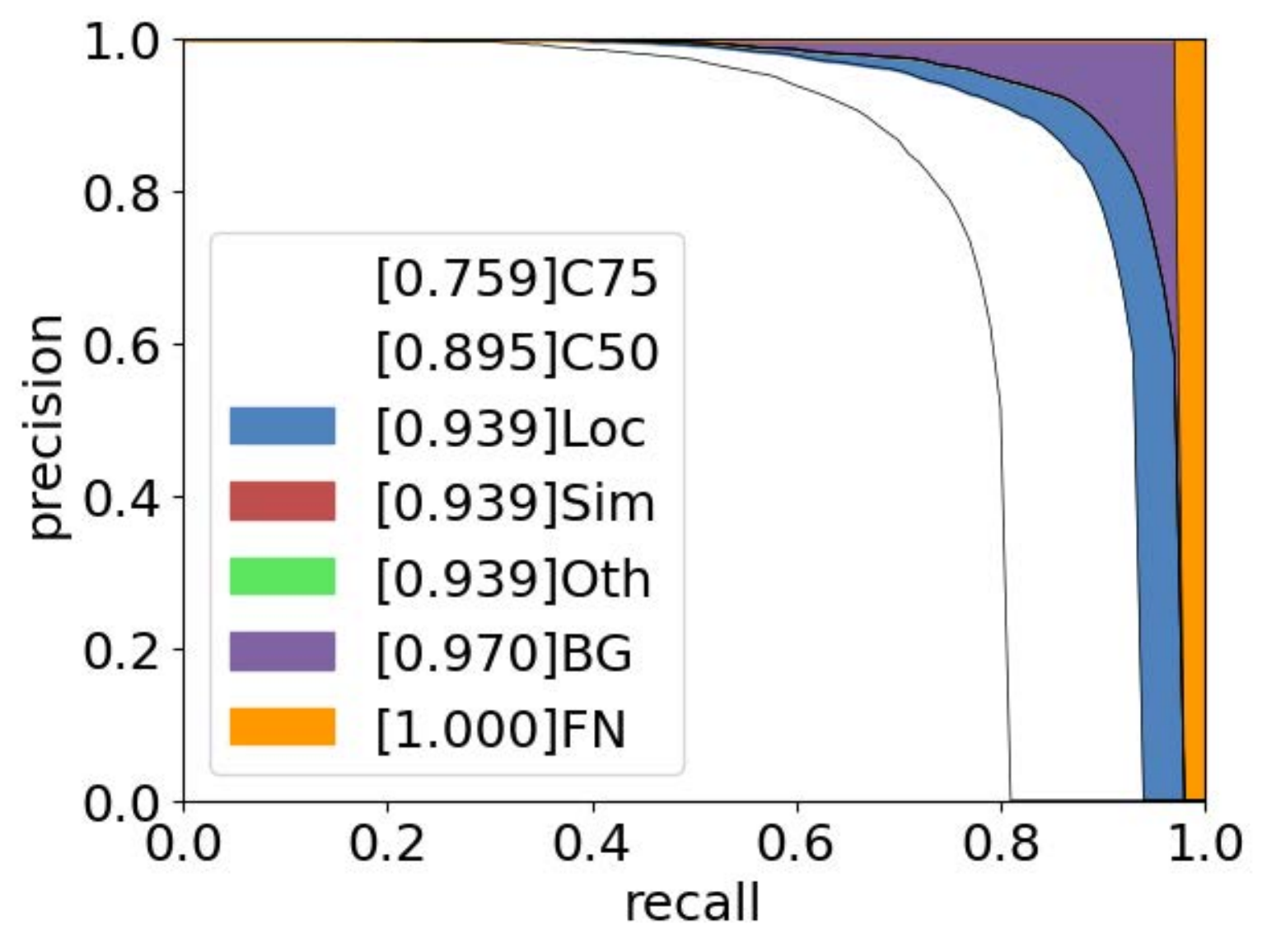}}}\hfill
\subfloat[\centering MoCo ]{{\includegraphics[width=0.2\textwidth]{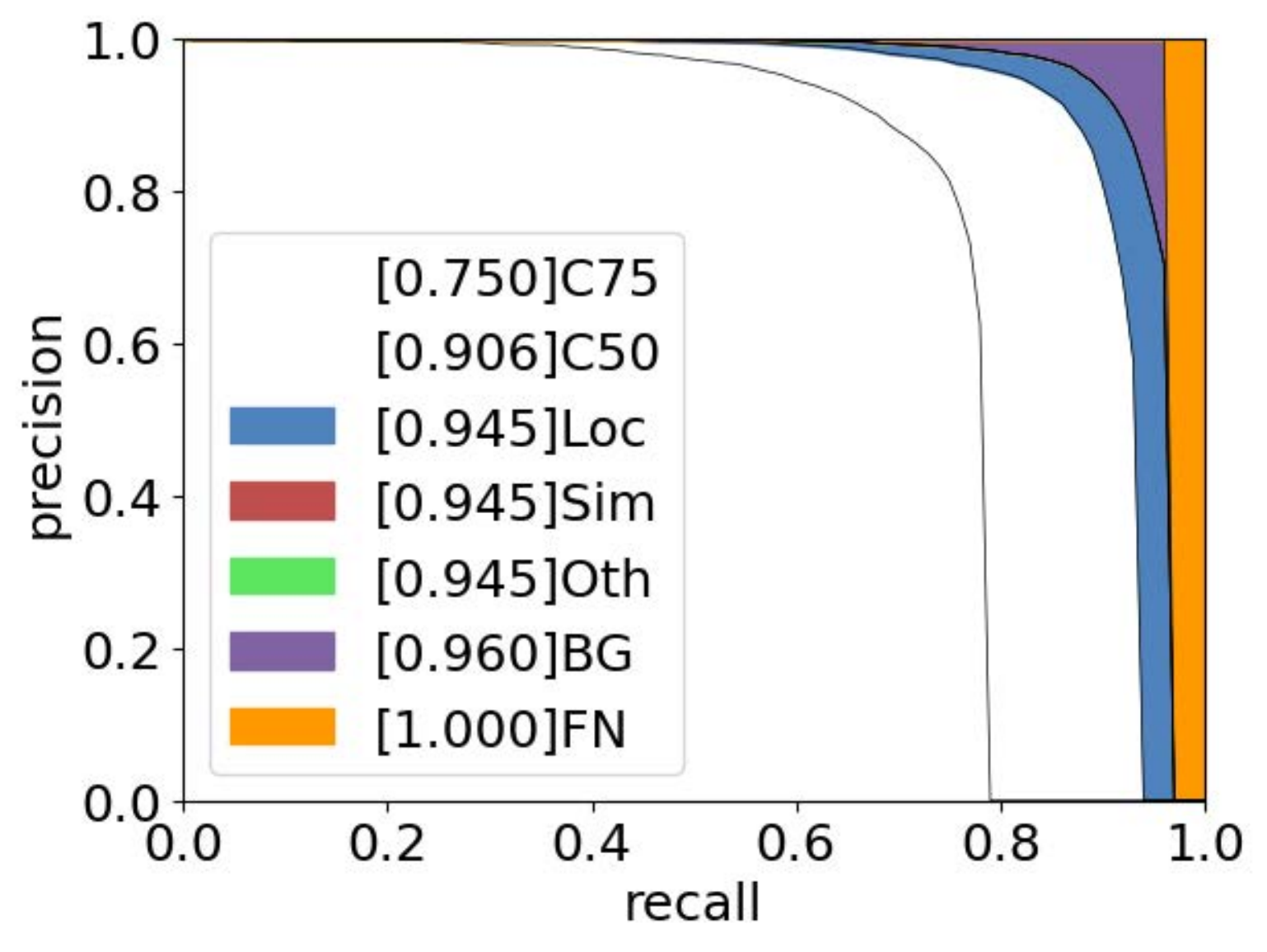}}}\hfill
\subfloat[\centering SimMIM-ViT ]{{\includegraphics[width=0.2\textwidth]{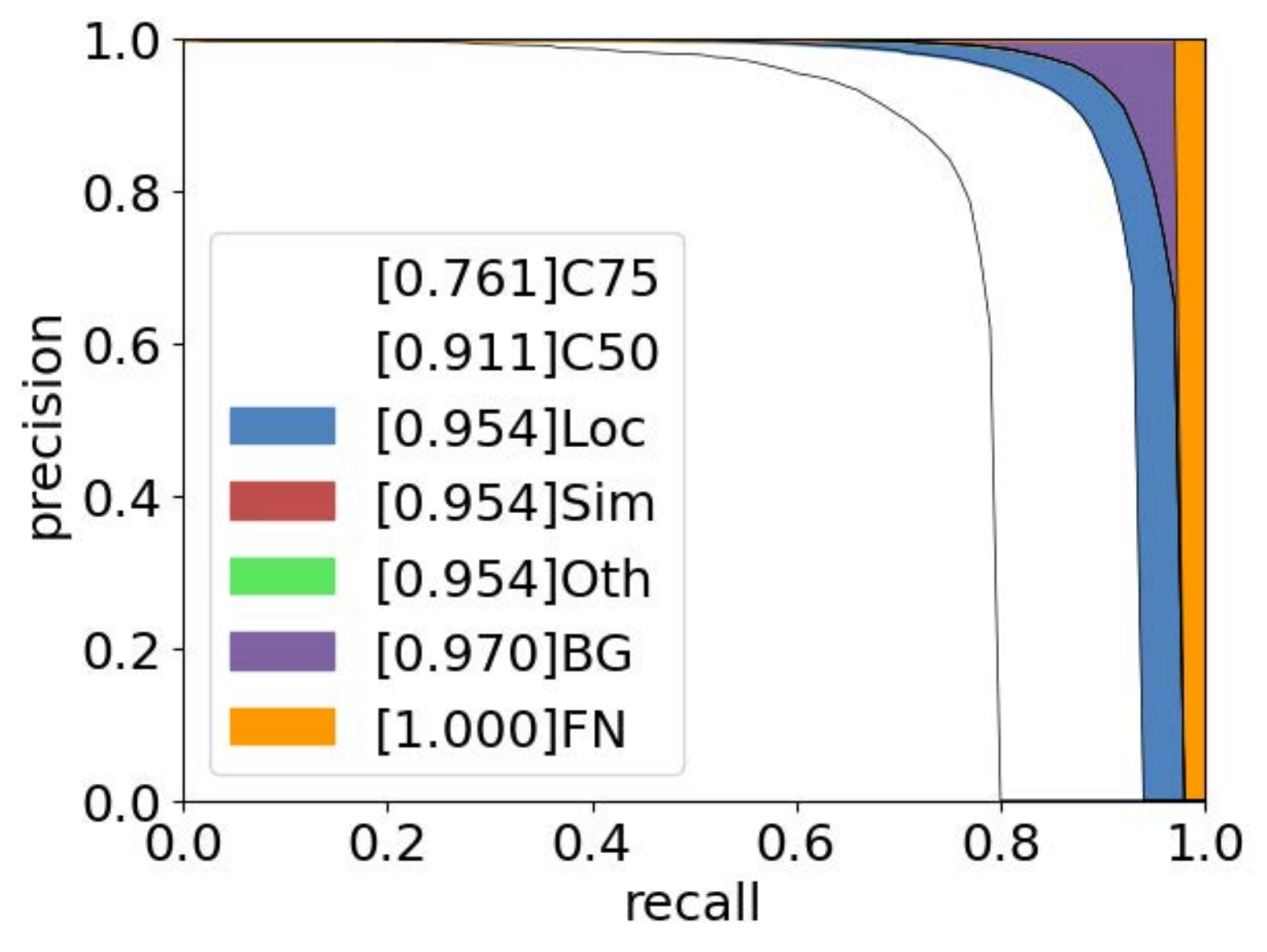}}}\hfill
\subfloat[\centering SimMIM-Swin ]{{\includegraphics[width=0.2\textwidth]{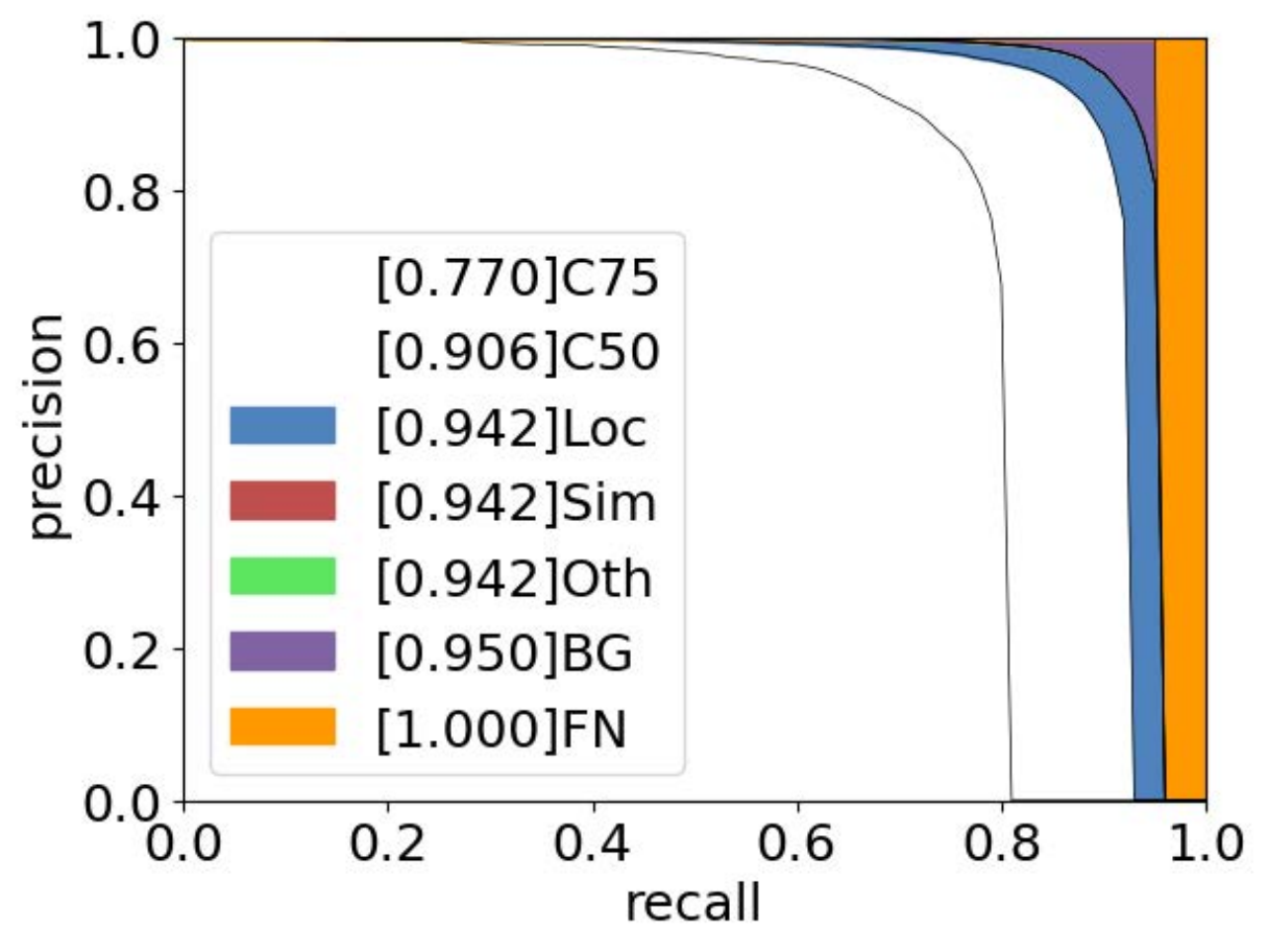}}}
\caption{(a-e) Segmentation mask Precision-Recall curve at different IoU thresholds and (f-j) Segmentation mask error analysis of DAMA and other SSL methods in 7-channel.} 
\label{fig:segprecall}
\end{figure*}

\textcolor{nblue}{\textit{Cell segmentation.}} Pretrained and finetuned with the same ViT backbone architecture, except for SimMIM-Swin, and object detector frameworks Mask R-CNN \cite{maskrcnn}, our DAMA achieves the best performances compared to other SSL methods, see Table \ref{tab:segment} and Fig. \ref{fig:vizsegsample}, \ref{fig:segprecall}. DAMA significantly outperforms all baseline methods. We hypothesize that DAMA is more capable of utilizing the under-represented regions around cells to resolve ambiguous cases where cells are dense and overlapping. Two observations can support this hypothesis. First, MoCo-v3's \cite{mocov3} aims to increase the mutual information of two augmented views coming from the same image, e.g., $I(X_1,X_2)$. The random augmentation operations encourage MoCo-v3 to focus on the cell body while discarding the other features since they are likely to decrease the mutual information. Second, MAE \cite{mae} and SimMIM \cite{simmim} have a similar learning approach that learns to reconstruct missing image patches from uncorrupted patches, e.g., $I(X_{mask}, X_{unmask})$. However, the high random masking ratios in these two methods (e.g., 0.8) substantially reduce the model's ability to focus on under-represented regions.  
In addition, we also experiment with SimMIM with Swin \cite{swin} architecture which is more efficient for segmentation tasks than ViT \cite{simmim,swin}. In contrast, our DAMA model emphasizes the under-represented information by adaptive masking strategy in each iteration and achieves the best segmentation performance. Fig.~\ref{fig:vizsegsample} illustrates that DAMA can segment clusters of cells better than other methods. The results demonstrate DAMA's significant advantage in analyzing multiplexed brain cell data.

Fig. \ref{fig:segprecall} shows the segmentation performances of DAMA and other methods. For DAMA, the overall average precision (AP) at IoU@.75 is 0.77. When localization errors \textit{Loc}\footnote{The meaning of errors, \textit{Loc, Sim, Oth, BG, etc.,} are better explained on the COCO dataset website \url{https://cocodataset.org/\#detection-eval}}. are ignored, the precision-recall (PR) curve at IoU@.1, the AP increased to 0.956. Since we only segment the cell body from the background regardless of the cell type, e.g., having only one class, class-related error metrics, \textit{Sim} and \textit{Oth}, are unchanged. Removing background false positives \textit{BG} would increase the performance by 0.014 (to 0.97 AP), and the rest of the errors are missed detections. In summary, DAMA's errors come from imperfect localization \textit{Loc} and confusing background \textit{BG}. Regarding other methods, the amount of \textit{Loc} and \textit{BG} errors are higher than those of DAMA.

\begin{table}[] 
\centering
\footnotesize
\begin{tabular}{l@{\hskip0.2cm}c@{\hskip 0.3cm}c@{\hskip 0.3cm}c@{\hskip 0.3cm}}
\hline
\multicolumn{1}{c}{\scriptsize Methods}         & \scriptsize Classification(\%)        & \scriptsize Detection(mAP)                        & \scriptsize Segmentation(mAP)\\
\hline
ViT random init.  & 89.93 $\pm$0.90 & 0.536 $\pm$0.006 & 0.559 $\pm$0.005 \\
Swin random init. & - & 0.495 $\pm$0.005                    & 0.513 $\pm$0.004 \\
MoCo-v3 \cite{mocov3} & 93.73 $\pm$0.45               & 0.530 $\pm$0.007                    & 0.548  $\pm$0.007 \\
MAE  \cite{mae} & 91.82 $\pm$0.97                  & 0.570 $\pm$0.006                     & 0.586  $\pm$0.008 \\
SimMIM-ViT \cite{simmim} & -             & 0.549 $\pm$0.006                     & 0.570  $\pm$0.006 \\
SimMIM-Swin \cite{simmim} & -           & 0.576 $\pm$0.004                    & 0.596  $\pm$0.004 \\
DAMA (ours)  & 92.88 $\pm$0.89                 & 0.572 $\pm$0.005                      & 0.590 $\pm$0.006 \\
\hline
\end{tabular}
\caption{Data efficiency comparison in terms of the mean and standard deviation of using 10\% of training data for finetuning on the classification and detection/segmentation tasks. 
}
\label{tab:effclsseg}
\vspace{-10pt}
\end{table}

\begin{figure}[]
\centering
\includegraphics[width=0.42\textwidth,keepaspectratio]{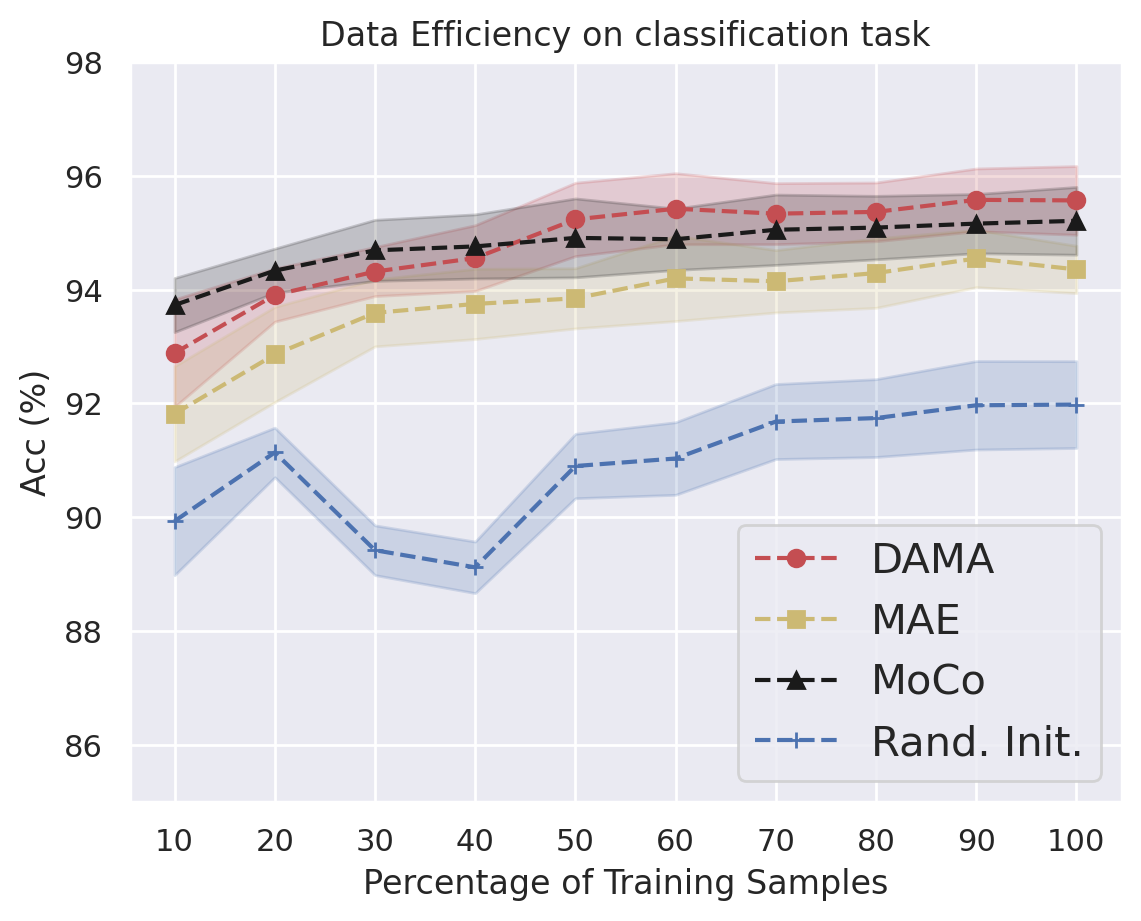}
\caption{Data efficiency plot in terms of the mean (center line) and standard deviation (colored band around the center line) when using 10\%-100\% of training data for finetuning on the cell classification task.}
\label{fig:dataeff}
\vspace{-10pt}
\end{figure}

\textcolor{nblue}{\textit{Data efficiency.}} Table \ref{tab:effclsseg} shows the data efficiency comparison of using 10\% of training data for
finetuning in the classification and detection/segmentation task in terms of mean and standard deviation. Similarly, Fig. \ref{fig:dataeff} presents the data efficiency comparison of using 10\%-100\% of training data for finetuning on the classification task. DAMA achieves consistently high data efficiency in both classification and segmentation tasks. The data efficiency experiments also demonstrate the advantages of SSL against random initialization. Throughout these experiments, we used the same validation sets used for experiments in Table \ref{tab:sota} and \ref{tab:segment}, respectively.

In the classification task, we repeatedly use 10\%-100\% of training data, i.e., 80-800 cell images per class, for each 10-fold finetuning, resulting in 100 experiments for each method. We also re-used the pre-trained weight on \textit{Set III} for classification experiments. Our DAMA and MoCo-v3 \cite{mocov3} achieve the best similar result compared to MAE \cite{mae} and randomly initialized weight. However, MoCo-v3 has slightly better performances when using 10\%-40\% of the original training data. This is due to its contrastive learning principle that benefits classification tasks. Unfortunately, the contrastive objective seems to hurt MoCo-v3 performances for detection and segmentation tasks as discussed below.

Regarding the segmentation task, due to the time consumption for each experiment, we only used 10\% of training data for finetuning by randomly splitting the data into ten (10) folds and using each fold for training, resulting in a total of 10 experiments for each method. In this task, MoCo-v3 has the worst performance among SSL methods. MAE \cite{mae} and SimMIM \cite{simmim} have a similar learning approach that learns to reconstruct missing image patches from uncorrupted patches. Hence, they might learn useful features for segmentation and not only concentrate on the cell body as MoCo-v3. Using the same backbone architecture ViT \cite{vit}, our DAMA achieves better performance by employing the adaptive sampling strategy and emphasizing the under-represented information regardless of the high masking ratio, unlike MAE and SimMIM-ViT. SimMIM-Swin has the best segmentation performance by using the Swin \cite{swin} backbones architecture which is more efficient for segmentation tasks than ViT \cite{swin,simmim}. As a comparison, SimMIM-ViT's result is significantly lower than SimMIM-Swin and DAMA.

\subsubsection{ImageNet-1k} To demonstrate the potential of DAMA on other natural image types, we present DAMA's result on ImageNet-1k \cite{imagenet} in Table \ref{tab:resultimagenet}. DAMA is competitive with other state-of-the-art algorithms despite smaller numbers of pre-trained epochs and without any ablation experiment for searching optimal hyper-parameters. Due to the computational resource needed for training on such a large-scale dataset, we perform only a \textit{single} pretraining/finetuning experiment on ImageNet-1k with the same configuration as for training on the brain image dataset, except for the image size and pretraining/finetuning batch size as $224\times224\times3$, and $4096/1024$, respectively.

\begin{table}[t]
\centering
\footnotesize
\begin{tabular}{lcc}
\hline
Methods     & Pretrained Epochs & Acc (\%) \\
\hline
MoCo-v3 \cite{mocov3}    & 600               & 83.2     \\
BEiT \cite{beit}       & 800               & 83.4     \\
SimMIM \cite{simmim}     & 800               & 83.8     \\
Data2Vec \cite{data2vec}   & 800               & 84.2     \\
DINO \cite{dino}       & 1600              & 83.6     \\
iBOT \cite{zhou2021ibot}       & 1600              & 84.0     \\
MAE \cite{mae}       & 1600              & 83.6     \\
DAMA (ours) & 500               & 83.2    \\
\hline
\end{tabular}
\caption{Comparisons results of DAMA and state-of-the-arts on ImageNet-1k.\label{tab:resultimagenet}}
\end{table}
\begin{figure}[t]
\centering
\includegraphics[width=0.45\textwidth,keepaspectratio]{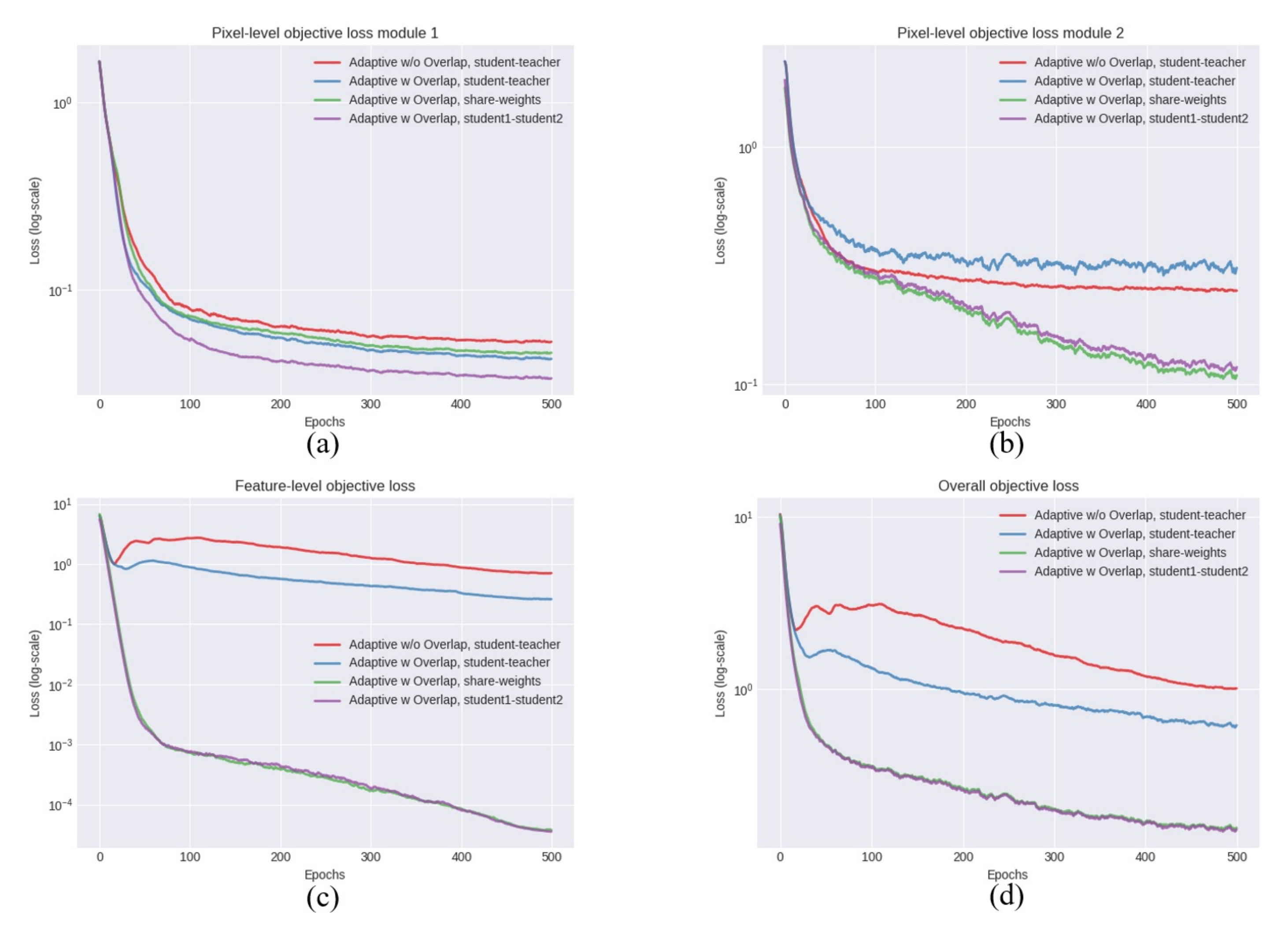}
\caption{Pre-training curves of different model settings on adaptive masking condition: Pixel-level reconstruction loss in (a) network1 and (b) network2, (c) feature-level regression loss, and (d) total loss.}
\label{fig:traincurve}
\vspace{-5pt}
\end{figure}

\subsection{Ablation Studies}\label{sec:ablation}
\begin{figure}[]
\centering
\includegraphics[width=0.47\textwidth,]{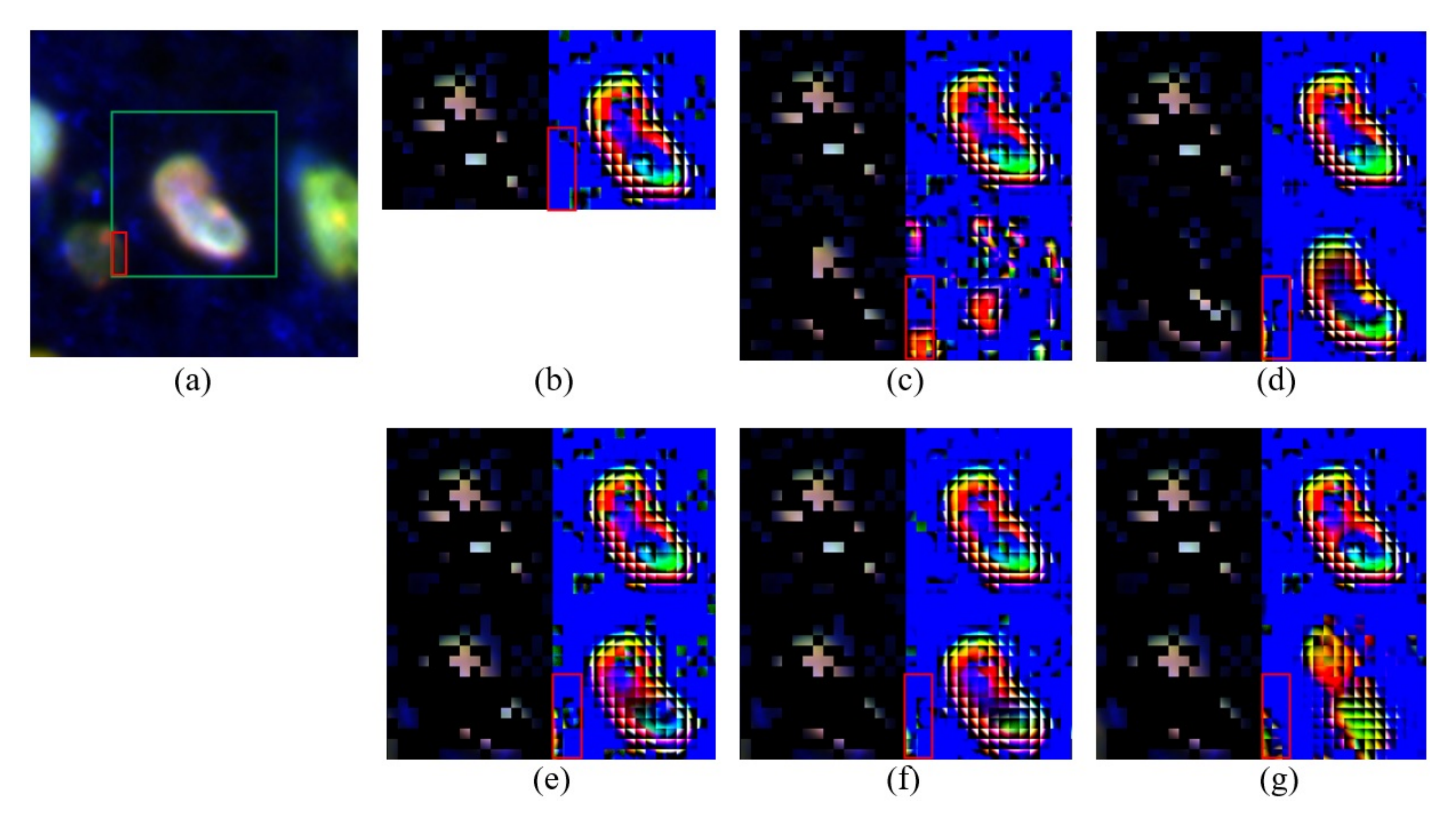}
\caption{Reconstruction effect of mask sampling strategy between MAE (b) and our (c-g) from (a) the original input. While MAE can not reconstruct the region in \textcolor{red}{red box} in a single iteration, our DAMA is able to do that with adaptive sampling. Images are converted to RGB 3 channels for visualization.}
\label{fig:recons}
\end{figure}

In Fig. \ref{fig:traincurve}, Fig. \ref{fig:recons}, and Table \ref{tab:abl}, we ablate our DAMA with different masking strategies, model strategies, and masking ratios. See the Appendix for more results.

\textcolor{nblue}{\textit{Masking Strategy and Masking Ratio.}} We compare our DAMA with \textit{adaptive} and \textit{random} masking strategies and analyze how they affect the finetuning results. We vary the masking ratio in the range of (60\%-90\%). The masking ratio of 80\% achieves a better result for all three masking settings. Similarly, MAE reports high masking ratios, ideally ranging in $60\%-80\%$ for good finetuning performance on ImageNet-1K. Primarily, adaptive masking produces better accuracy with different masking ratios in the same training condition. These experiments justify the effectiveness of our method. \textit{Overlapping}, a step in the proposed adaptive masking, means some unmasked patches in $X_1^m$ will also be unmasked in $X_2^m$. The overlapping ratio is set at 50\% for all experiments. No overlapping between unmasked input $X_1^m$ and $X_2^m$ leads to more challenging optimization of $\mathcal L_{f}$ as illustrated in Fig. \ref{fig:traincurve}. Specifically, the loss curves for none overlapping experiments are higher than others. These results demonstrate the effect of adaptive masking and feature-level regression on the overall objective function. 

\begin{table}[]
\centering
\footnotesize
\begin{tabular}{|l|c|cccc|}
\hline
\multirow{2}{*}{Masking strategy}      & \multirow{2}{*}{Model settings} & \multicolumn{4}{c|}{Masking ratio \%}                                                                                                  \\ \cline{3-6} 
                                      &                             & \multicolumn{1}{c|}{60}             & \multicolumn{1}{c|}{70}             & \multicolumn{1}{c|}{80}                   & 90             \\ \hline
Rand. w overlap                       & student-ema                 & \multicolumn{1}{c|}{94.65}          & \multicolumn{1}{c|}{94.67}          & \multicolumn{1}{c|}{\underline{94.76}}          & 94.65          \\ \hline
Adapt. w/o overlap               & student-ema                 & \multicolumn{1}{c|}{94.88}          & \multicolumn{1}{c|}{\underline{95.17}}    & \multicolumn{1}{c|}{95.05}                & 95.06          \\ \hline
\multirow{3}{*}{Adapt. w overlap} & student-ema                 & \multicolumn{1}{c|}{95.14}          & \multicolumn{1}{c|}{95.4}           & \multicolumn{1}{c|}{\underline{95.43}}          & 95             \\ \cline{2-6} 
                                      & shared weights              & \multicolumn{1}{c|}{94.81}          & \multicolumn{1}{c|}{\underline{95.13}}    & \multicolumn{1}{c|}{\underline{95.13}}          & 94.65          \\ \cline{2-6} 
                                      & student1-student2           & \multicolumn{1}{c|}{\textbf{95.34}} & \multicolumn{1}{c|}{\textbf{95.46}} & \multicolumn{1}{c|}{\underline{\textbf{95.47}}} & \textbf{95.38} \\ \hline
\end{tabular}
\caption{Effect of sampling strategies, model settings, and masking ratios in classification accuracy. \textbf{Bold} are the highest score for each column, while \underline{underlined} are highest score for each row.\label{tab:abl}}
\vspace{-5pt}
\end{table}

\textcolor{nblue}{\textit{Student and Teacher Configurations.}}\label{subsec:modelsetting} We perform ablation studies with different variants of student and teacher network configurations. Specifically, we compare three configurations: student-teacher, shared-weights (student1-student1), and student1-student2. \textit{student-teacher} network is an exponential moving average (EMA) on the student weights \cite{moco}, and the update rule is $\theta_t\leftarrow \lambda\theta_t + (1-\lambda)\theta_s$, where $\lambda$ follows cosine scheduled from 0.996 to 1 during training \cite{byol,moco,dino}. Both networks use the same set of parameters in the \textit{shared-weights} setting. The \textit{student1-student2} setting consists of two independent networks, allowing more \textit{``freedom"} in optimizing objective functions. Table \ref{tab:abl} and Fig. \ref{fig:traincurve} show that the \textit{student-teacher} is less effective than \textit{shared-weights} (\textit{student1-student1}) and \textit{student1-student2} settings. In addition, utilizing an adaptive sampling strategy can improve the performance with the same network configuration. This supports the advantage of our method.

\textcolor{nblue}{\textit{Choice of $\alpha$.}} The choice of parameter $\alpha$ usually is case-by-case determined. Following \cite{morency,completer}, we keep the scale of feature-level regression loss $\mathcal L_{f}$ is $\frac{1}{10}$ to $\frac{1}{100}$ to the scale of the pixel-level reconstruction loss $\mathcal L_{p}$, see Fig. \ref{fig:traincurve}. We empirically found that setting $\alpha=1$ yields consistently good performance. The paper \cite{morency} shares a similar approach to our paper, except an additional contrastive loss, e.g., $\mathcal L_{total} = \alpha_1\:\mathcal L_{p} + \alpha_2\:\mathcal L_{f} + \alpha_3\:\mathcal L_{CL}$. They construct controlled experiments to study the effect of this sensitive and dataset-dependent hyper-parameter.

\textcolor{nblue}{\textit{Effect of Adaptive Masking Strategy.}}
We show in Fig. \ref{fig:recons} the reconstruction results of (b) MAE and (c-g) our five settings on the same setting as in Table \ref{tab:abl}. Except for (a), in (b-g), the upper row is the same random masking input $X_1^m$ (left) and its reconstruction result $G(Z_{X_1^m})$ (right), the lower row is the applied mask strategy input $X_2^m$ (left) and its reconstruction result $G(Z_{X_2^m})$ (right). From left to right, (a) original image is center cropped as input \textcolor{green}{green box}; (b) MAE's result, (c) our model-ema random sampling result; our adaptive sampling: (d) student-teacher without mask overlapping; (e) student-teacher with mask overlapping; (f) share-weights with mask overlapping; and (g) student1-student2 with mask overlapping. Without adaptive masking, (c) can not reconstruct properly. This could be explained as the teacher network being constrained by its student network and could not produce a reasonable inference with another random mask. In contrast, (d) also uses the student-teacher setting, but with the adaptive mask, can reconstruct the cell reasonably well. Note that, while MAE can not reconstruct the region in \textcolor{red}{red box} of Fig.~\ref{fig:recons}, which is part of another cell in a single iteration, our method can do that with adaptive sampling. This suggests MAE could leave out these fine-grain details even with several epochs in a random high masking ratio setting, i.e., 80\%. Conversely, our DAMA combines pixel- and feature-level optimization with adaptive masking to identify those details in every iteration with a high masking ratio. This supports the advantage of our method.

\section{Conclusion}
This paper introduces DAMA, the first adaptive masking SSL, as a unified framework for learning effective representations from multiplexed immunofluorescence brain images. DAMA leverages a dual loss consisting of a pixel-level reconstruction and a feature-level regression. Extensive experiments show that DAMA performs competitively on cell classification, detection, and segmentation tasks. Our work demonstrates the importance of adaptive mask sampling and information-theoretic dual loss function in SSL.

{\small
\bibliographystyle{plain}
\bibliography{egbib}
}


\begin{figure*}[ht]
\centering
\includegraphics[width=\textwidth]{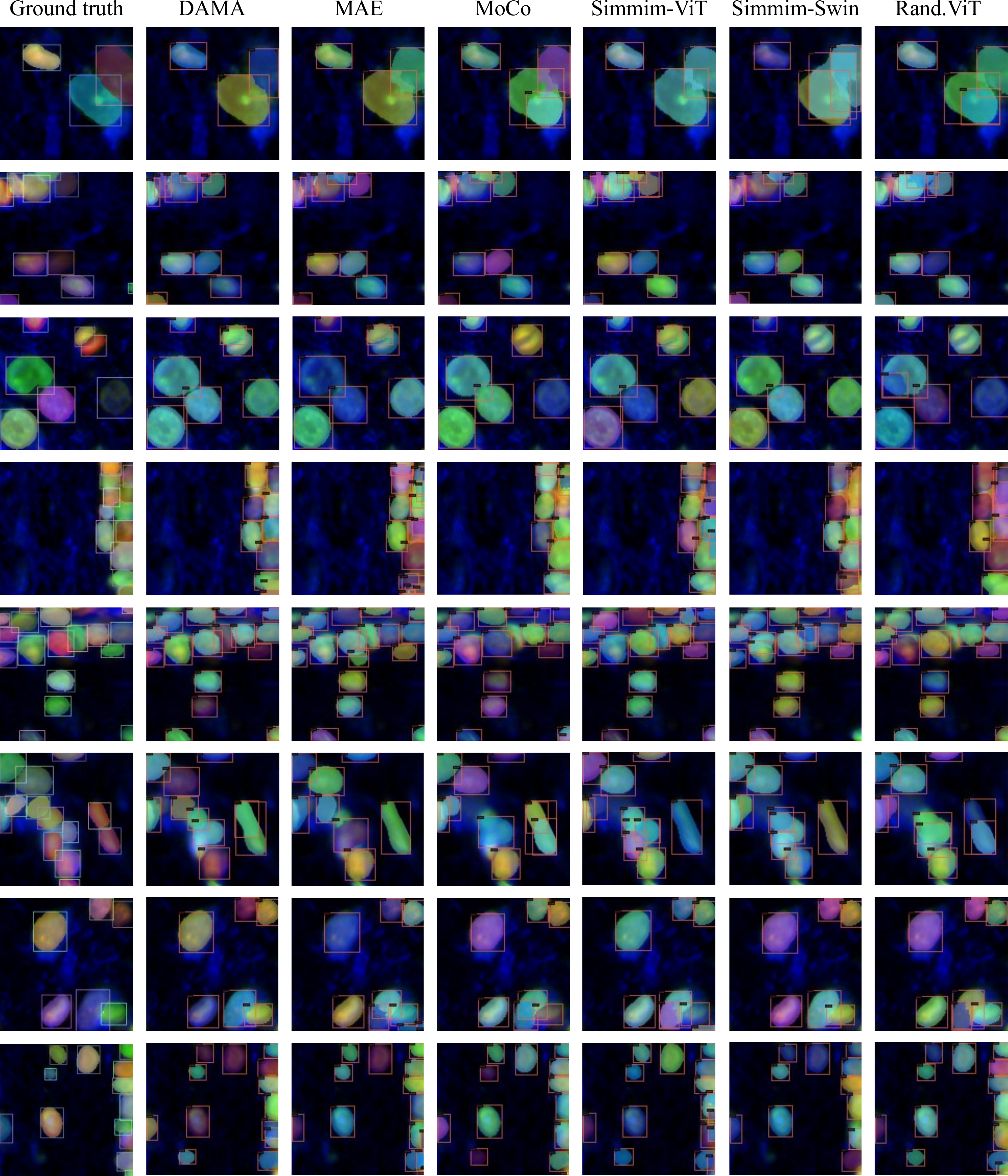}
\caption{Visualization segmentation validation set of DAMA and other methods at threshold IoU = 0.75. By focusing more on the contextual information, DAMA detects and segments cells better where cells are dense and overlap on each other, e.g., cluster of cell.}
\label{fig:vizseg}
\end{figure*}
\begin{figure*}[ht]
\centering
\subfloat[\centering DAMA ]{{\includegraphics[width=0.3\textwidth]{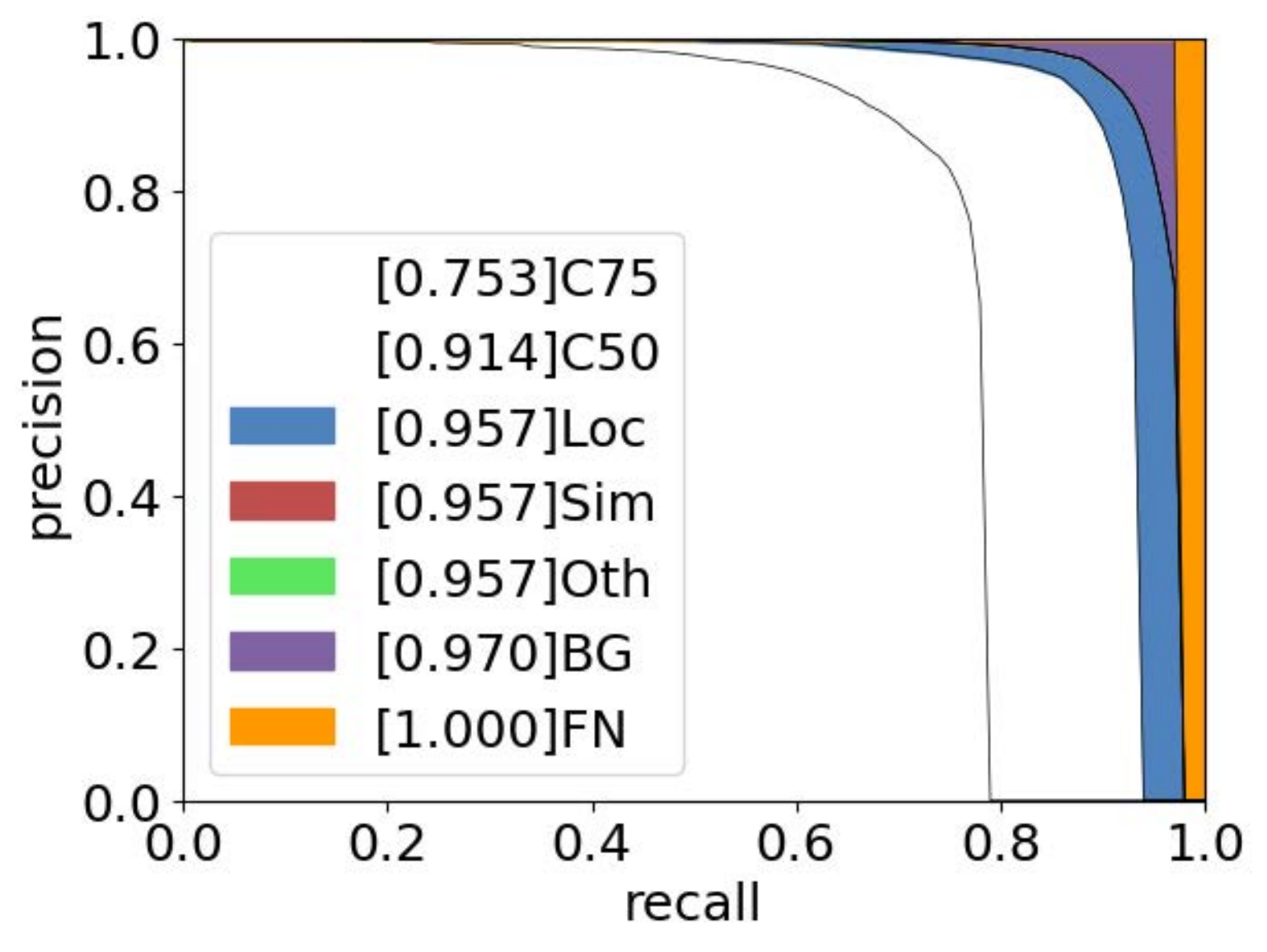}}}\hfill
\subfloat[\centering MAE ]{{\includegraphics[width=0.3\textwidth]{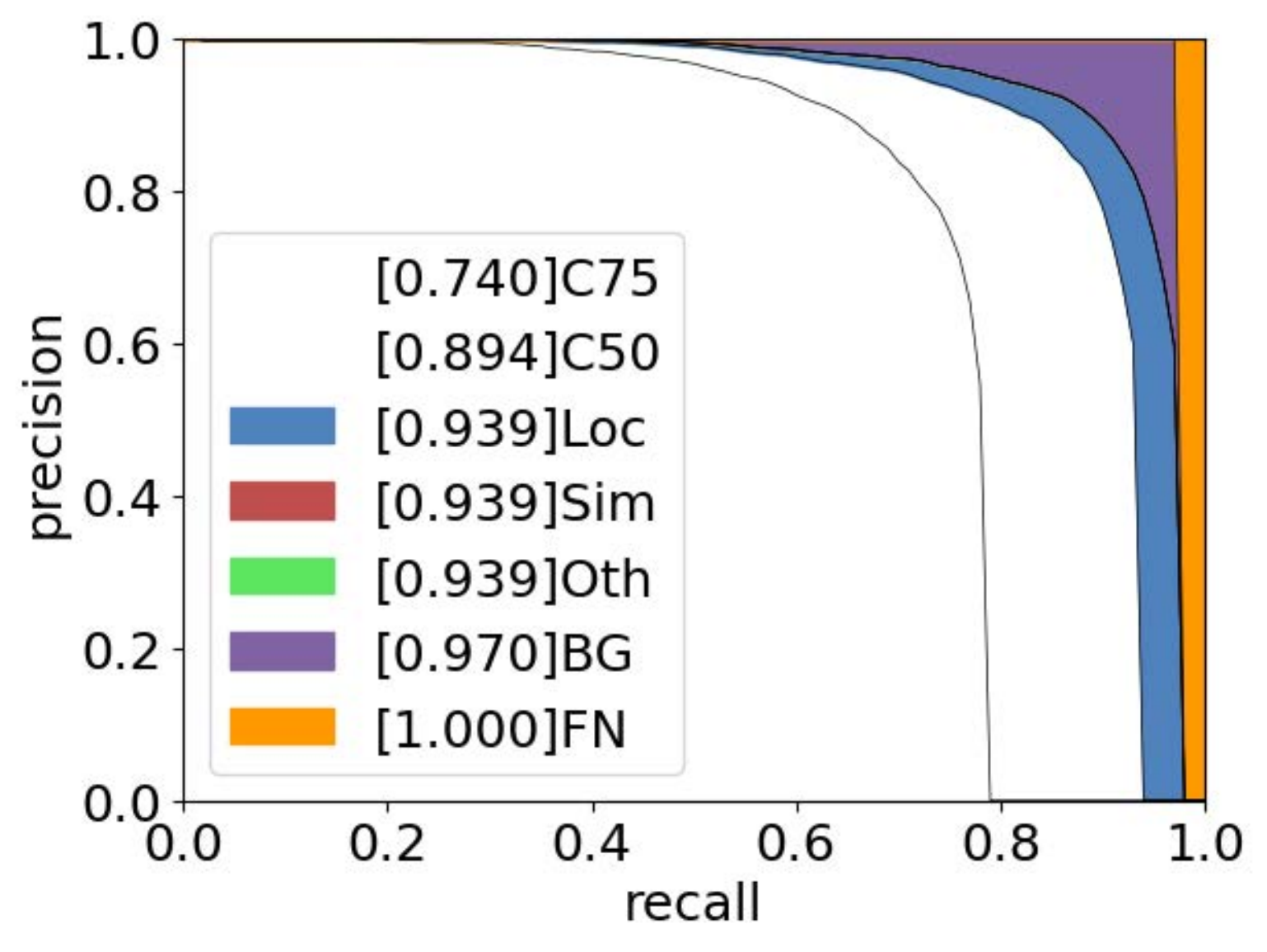}}}\hfill
\subfloat[\centering MoCo ]{{\includegraphics[width=0.3\textwidth]{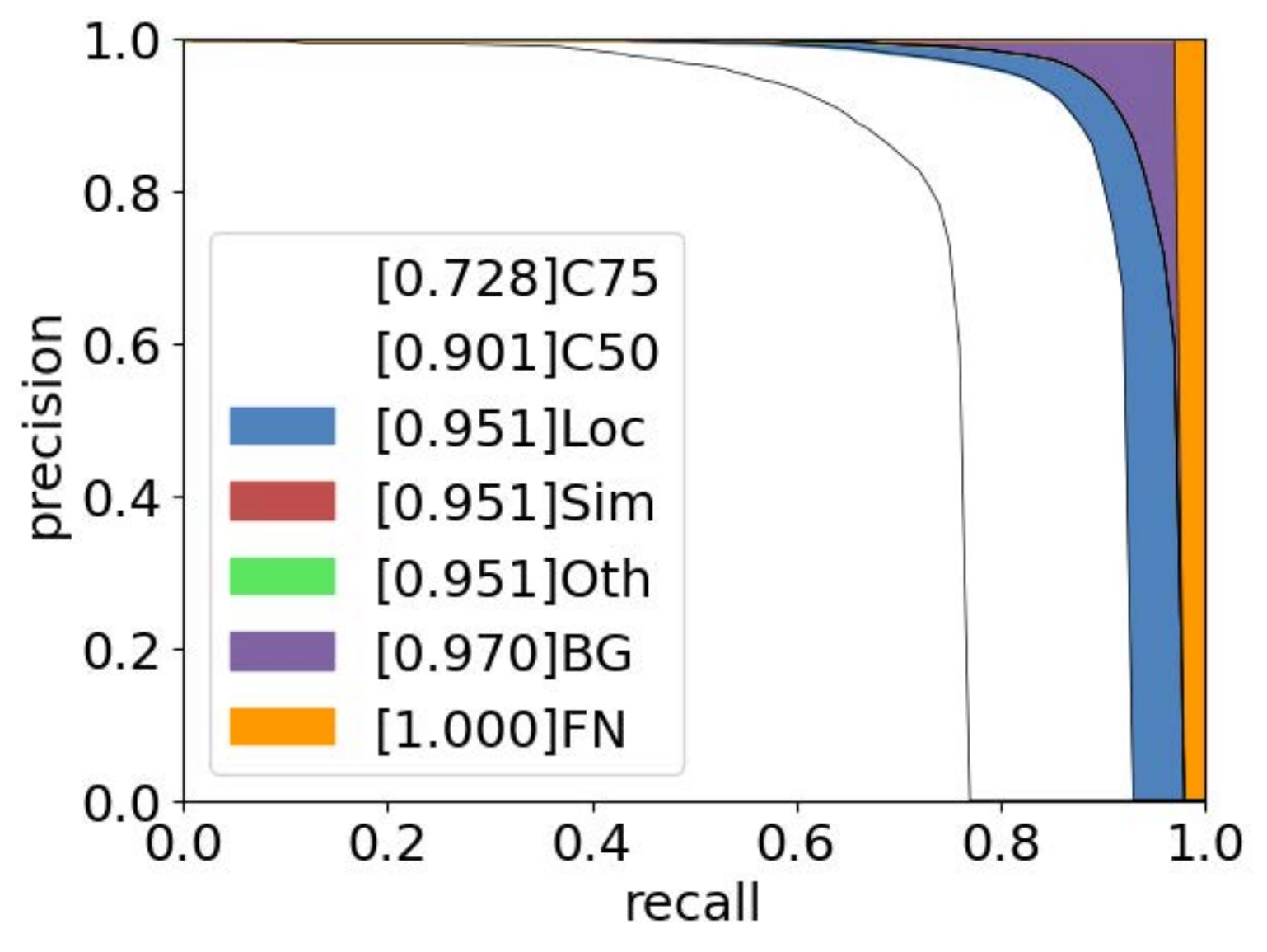}}} \\%
\subfloat[\centering SimMIM-ViT ]{{\includegraphics[width=0.3\textwidth]{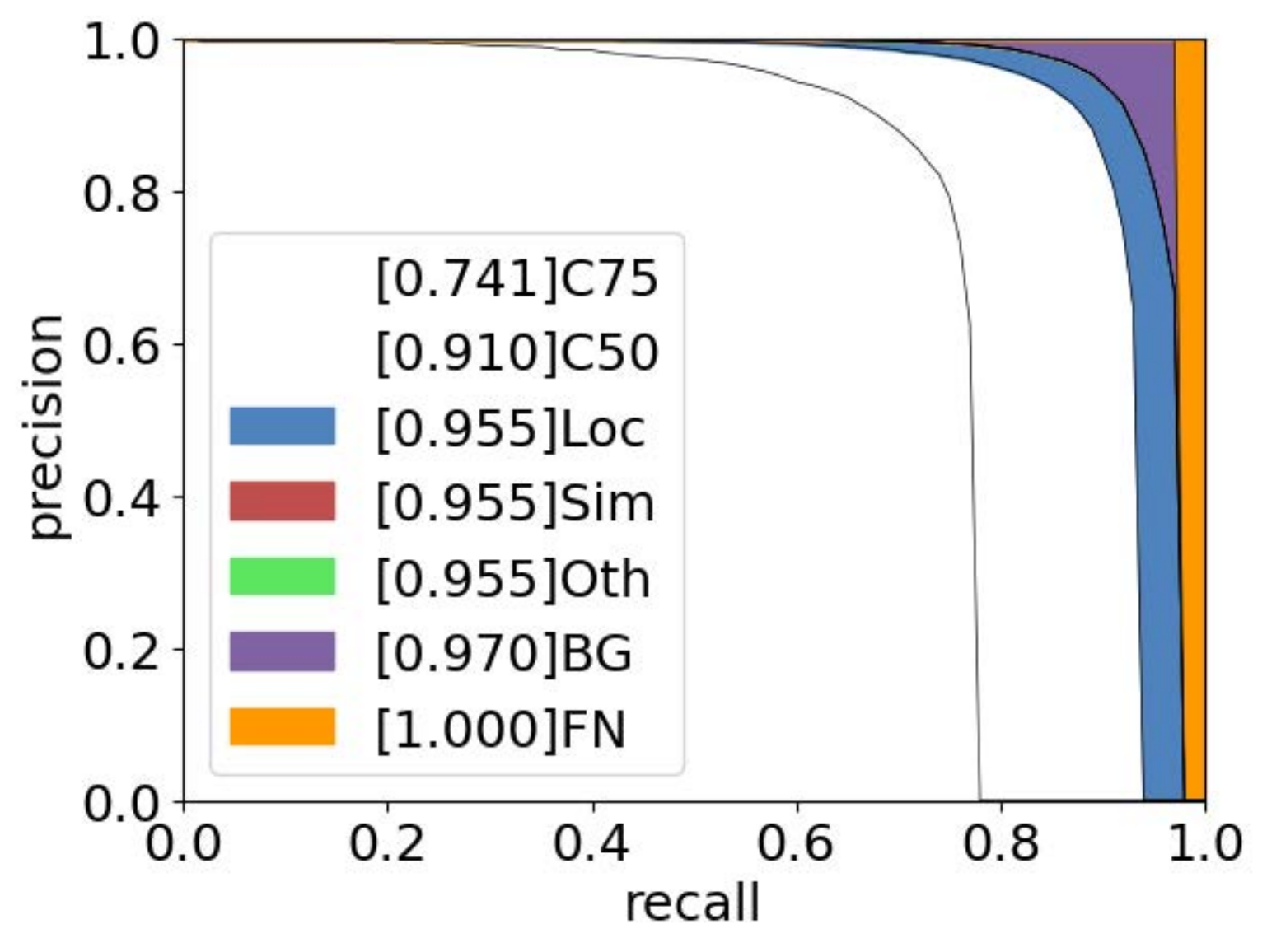}}}\hfill
\subfloat[\centering SimMIM-Swin ]{{\includegraphics[width=0.3\textwidth]{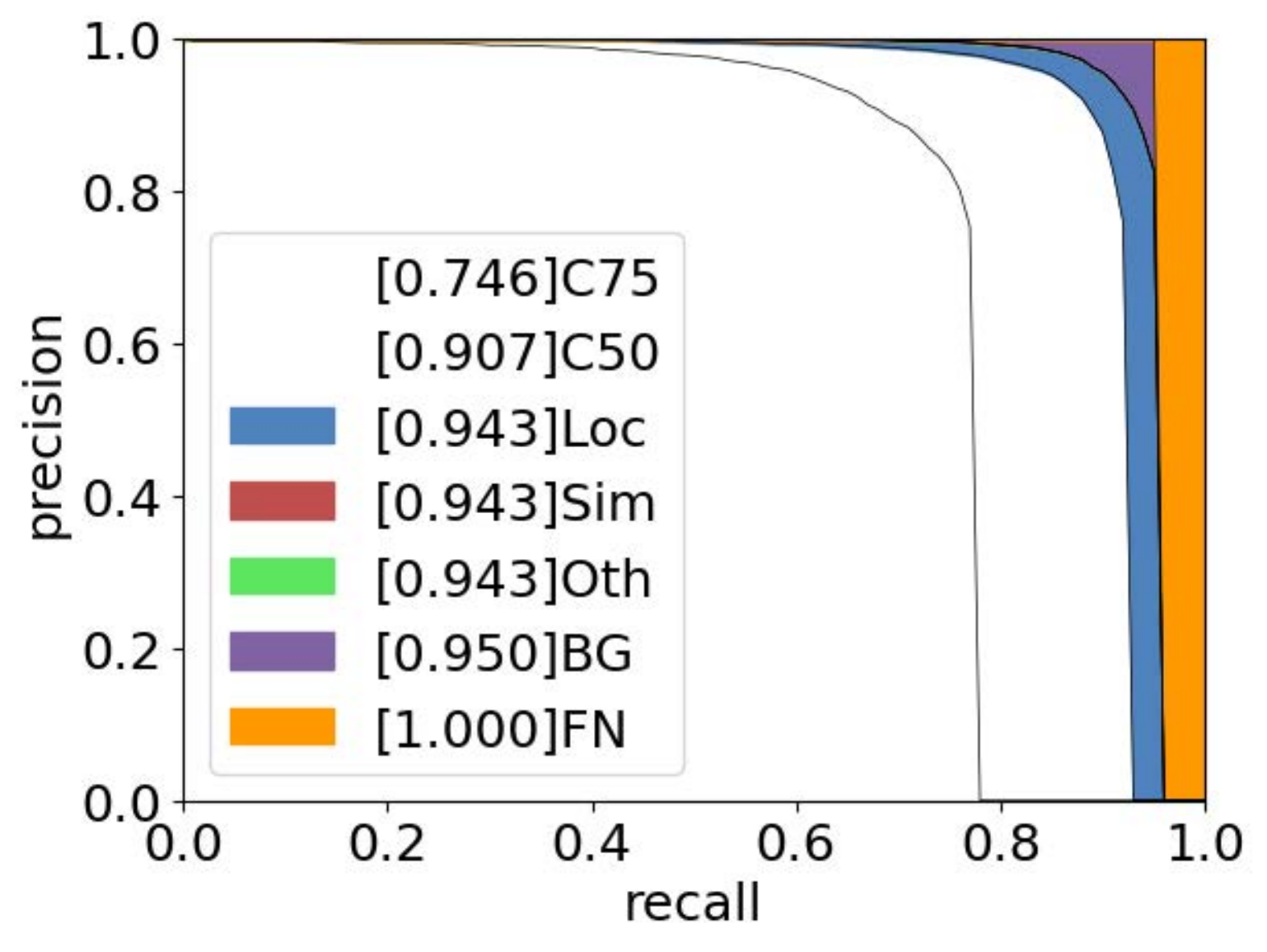}}}\hfill
\subfloat[\centering Random init. ViT ]{{\includegraphics[width=0.3\textwidth]{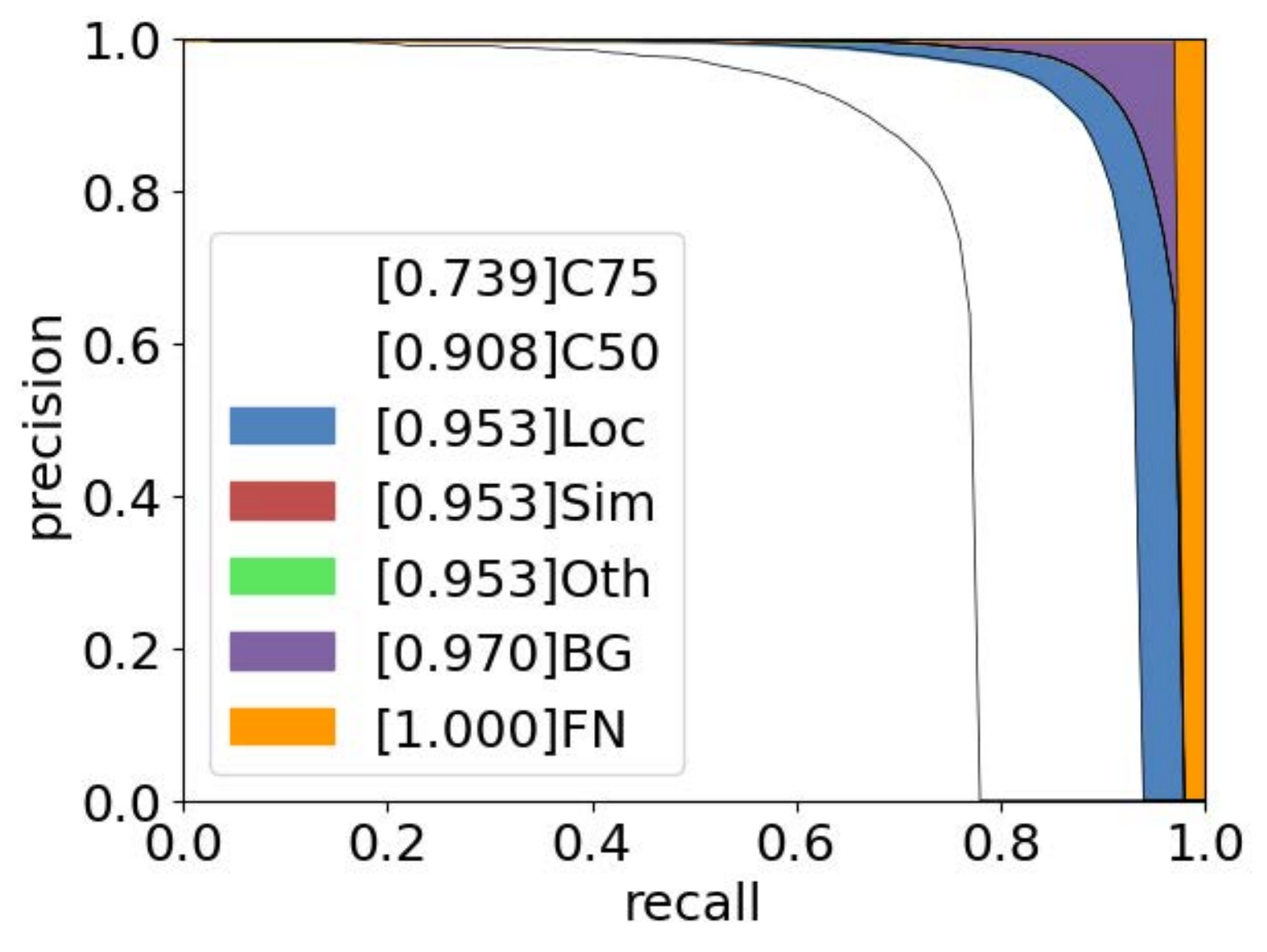}}}%
\caption{Bounding box overall-all-all Precision-Recall curve of DAMA and other SSL methods.} 
\label{fig:bboxprerecall}
\end{figure*}
\begin{figure*}[ht]
\centering
\subfloat[\centering DAMA ]{{\includegraphics[width=0.3\textwidth]{figs/segm/segm-allclass-allarea_dama10.pdf}}}\hfill
\subfloat[\centering MAE ]{{\includegraphics[width=0.3\textwidth]{figs/segm/segm-allclass-allarea_mae16.pdf}}}\hfill
\subfloat[\centering MoCo ]{{\includegraphics[width=0.3\textwidth]{figs/segm/segm-allclass-allarea_moco10.pdf}}} \\%
\subfloat[\centering SimMIM-ViT ]{{\includegraphics[width=0.3\textwidth]{figs/segm/segm-allclass-allarea_simmimvit.pdf}}}\hfill
\subfloat[\centering SimMIM-Swin ]{{\includegraphics[width=0.3\textwidth]{figs/segm/segm-allclass-allarea_simmimswin.pdf}}}\hfill
\subfloat[\centering Random init. ViT ]{{\includegraphics[width=0.3\textwidth]{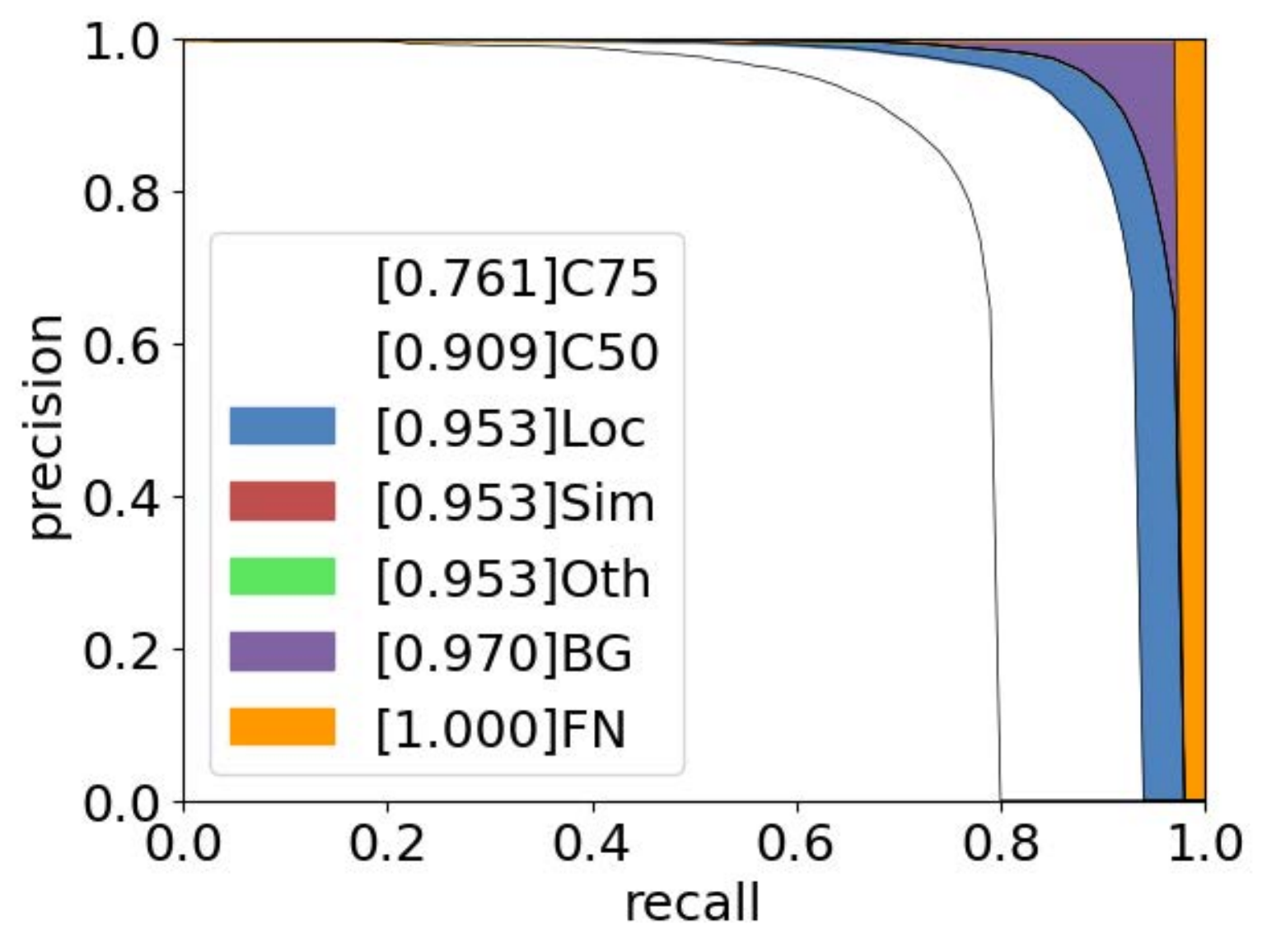}}}%
\caption{Segmentation mask overall-all-all Precision-Recall curve of DAMA and other SSL methods. Same as Fig. \ref{fig:segprecall}.} 
\label{fig:segprecall2}
\end{figure*}
\begin{figure*}[ht]
\centering
\subfloat[\centering DAMA ]{{\includegraphics[width=0.3\textwidth]{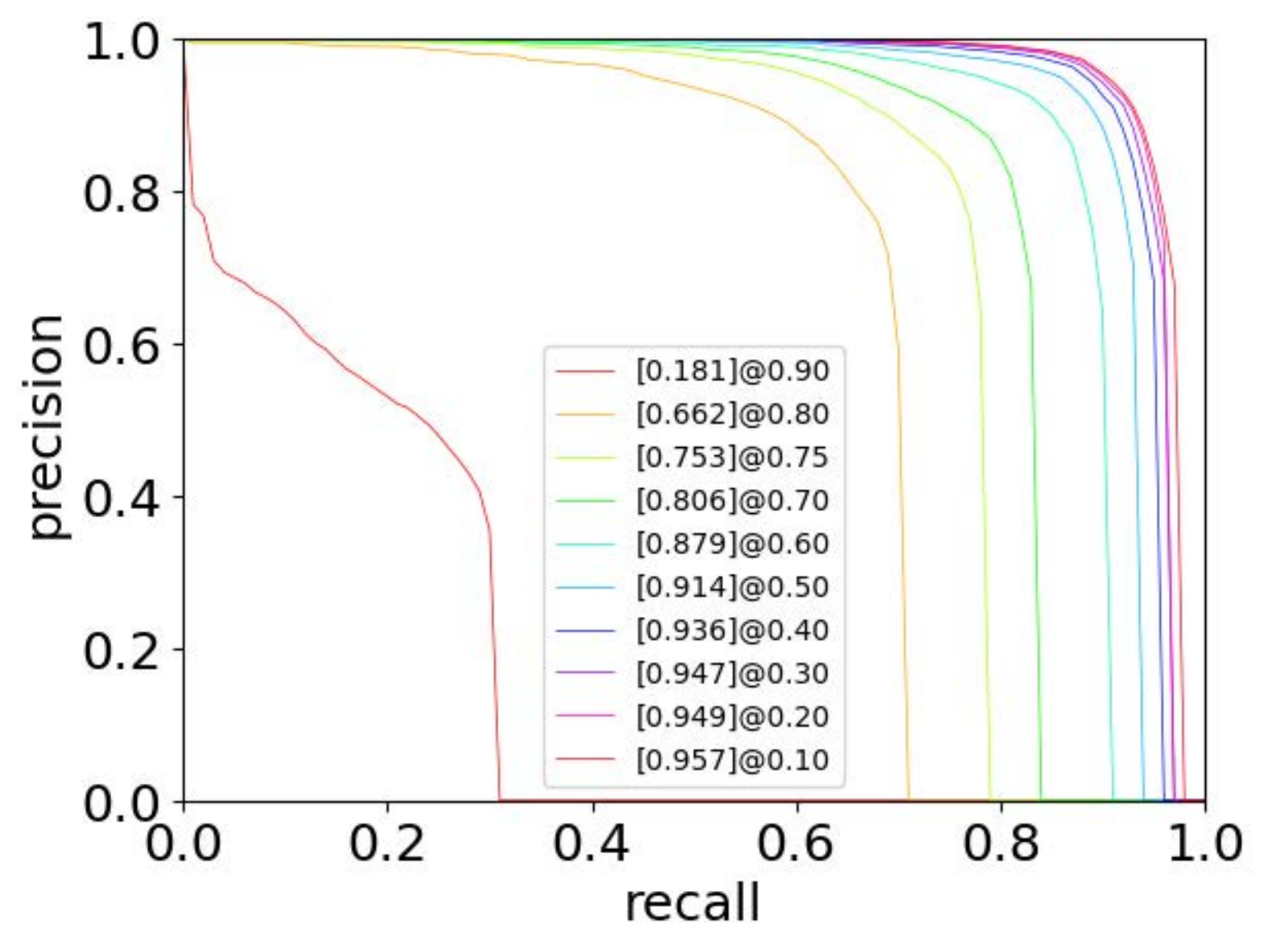}}}\hfill
\subfloat[\centering MAE ]{{\includegraphics[width=0.3\textwidth]{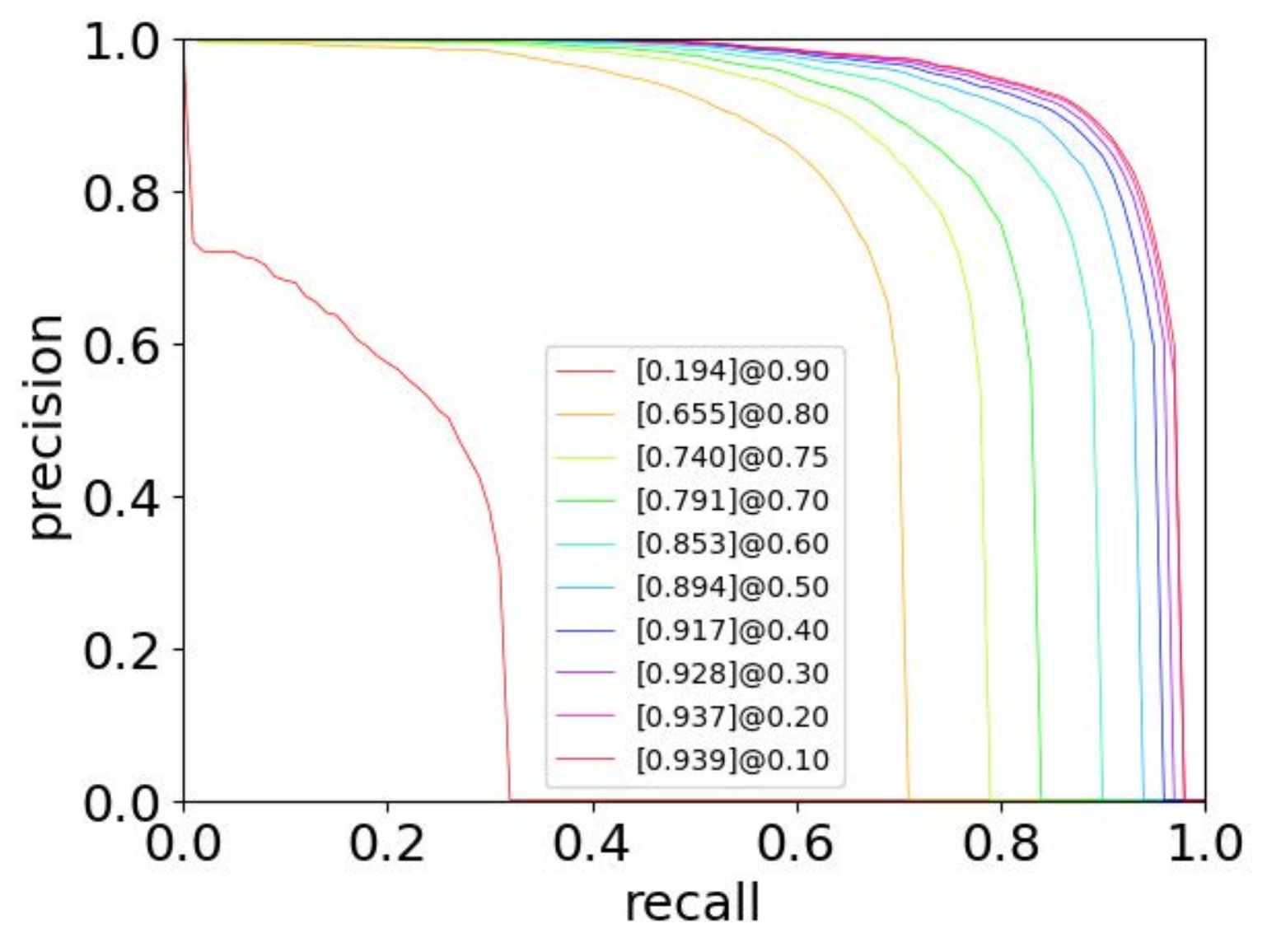}}}\hfill
\subfloat[\centering MoCo ]{{\includegraphics[width=0.3\textwidth]{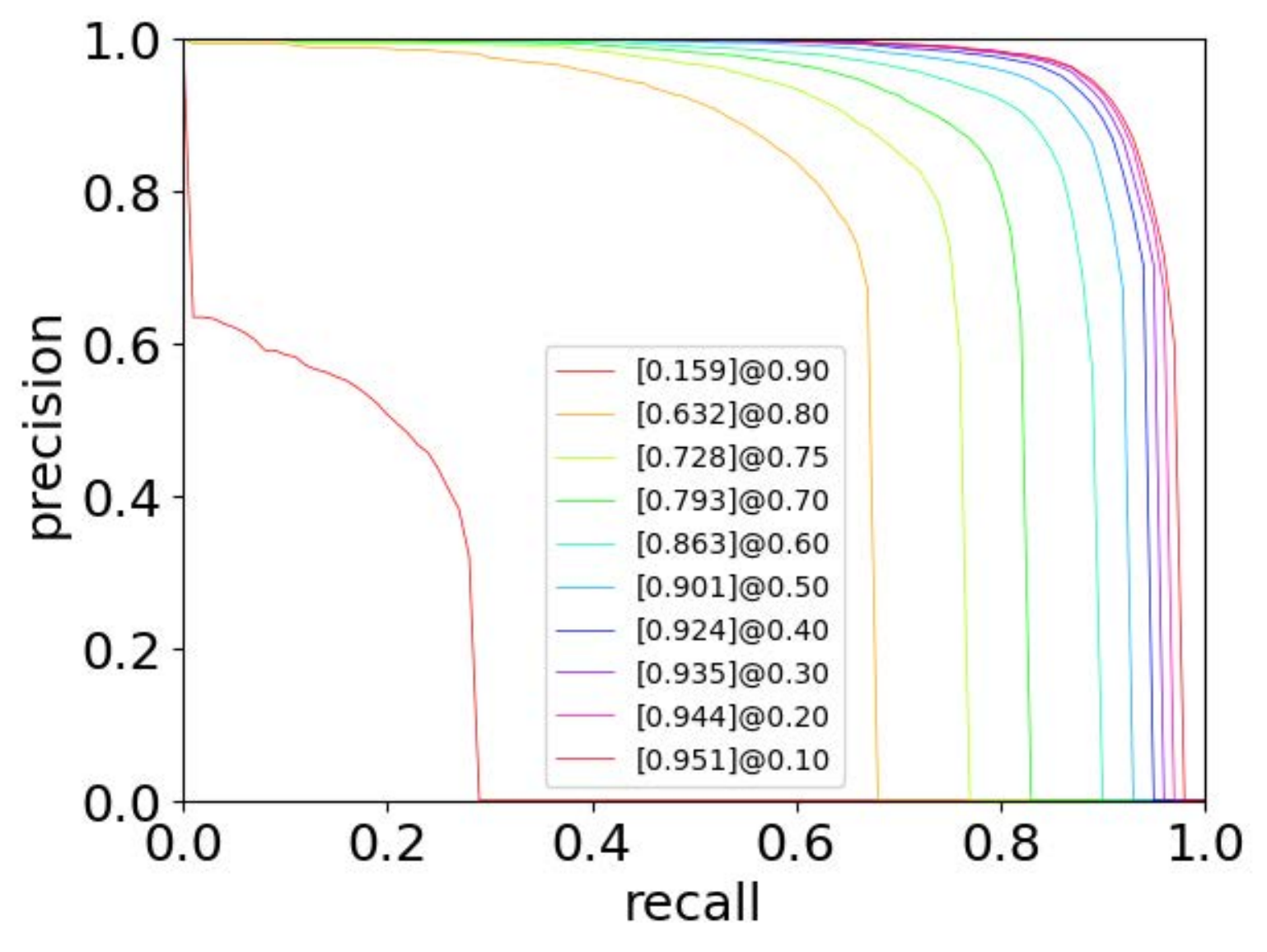}}} \\%
\subfloat[\centering SimMIM-ViT ]{{\includegraphics[width=0.3\textwidth]{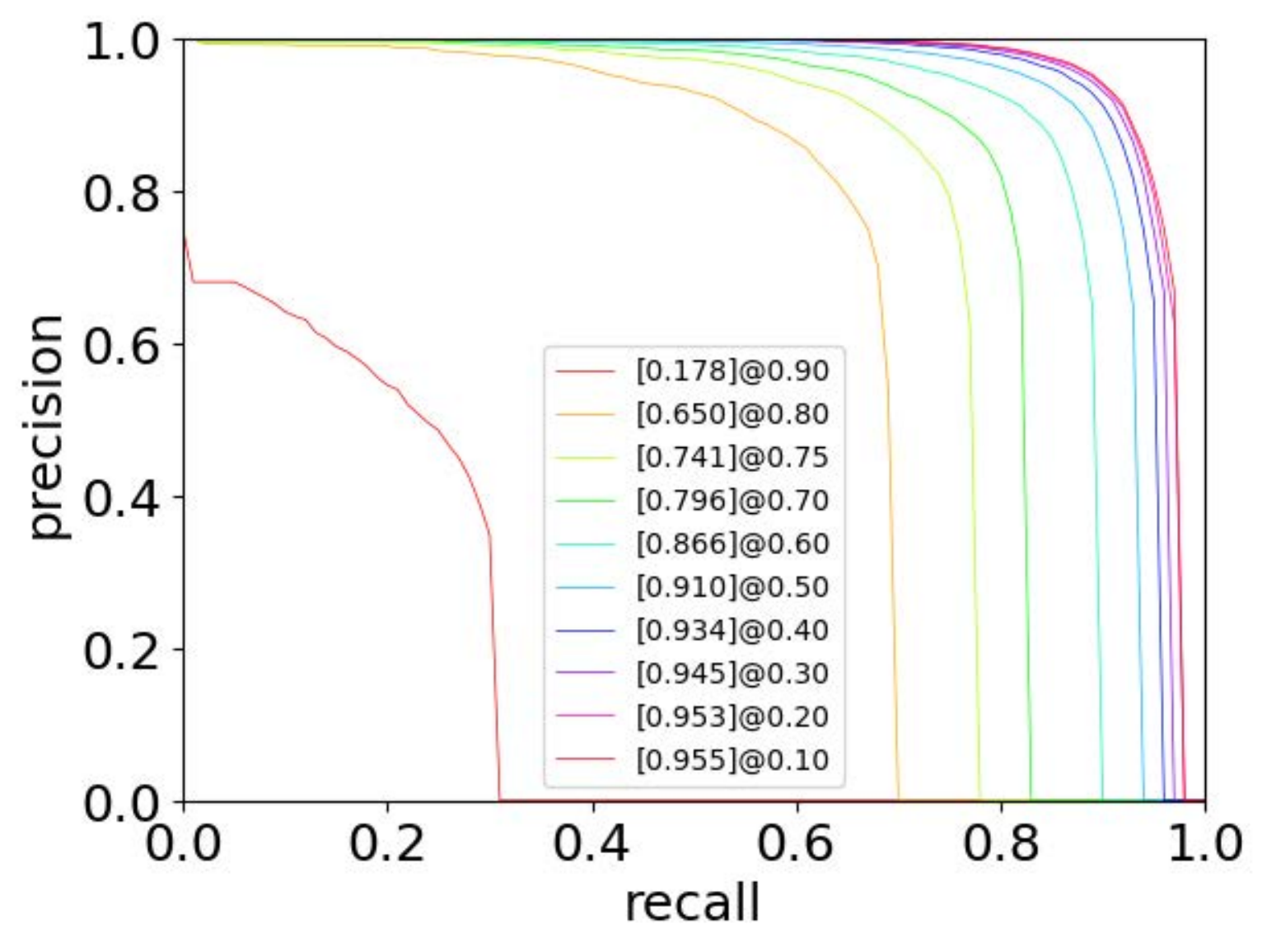}}}\hfill
\subfloat[\centering SimMIM-Swin ]{{\includegraphics[width=0.3\textwidth]{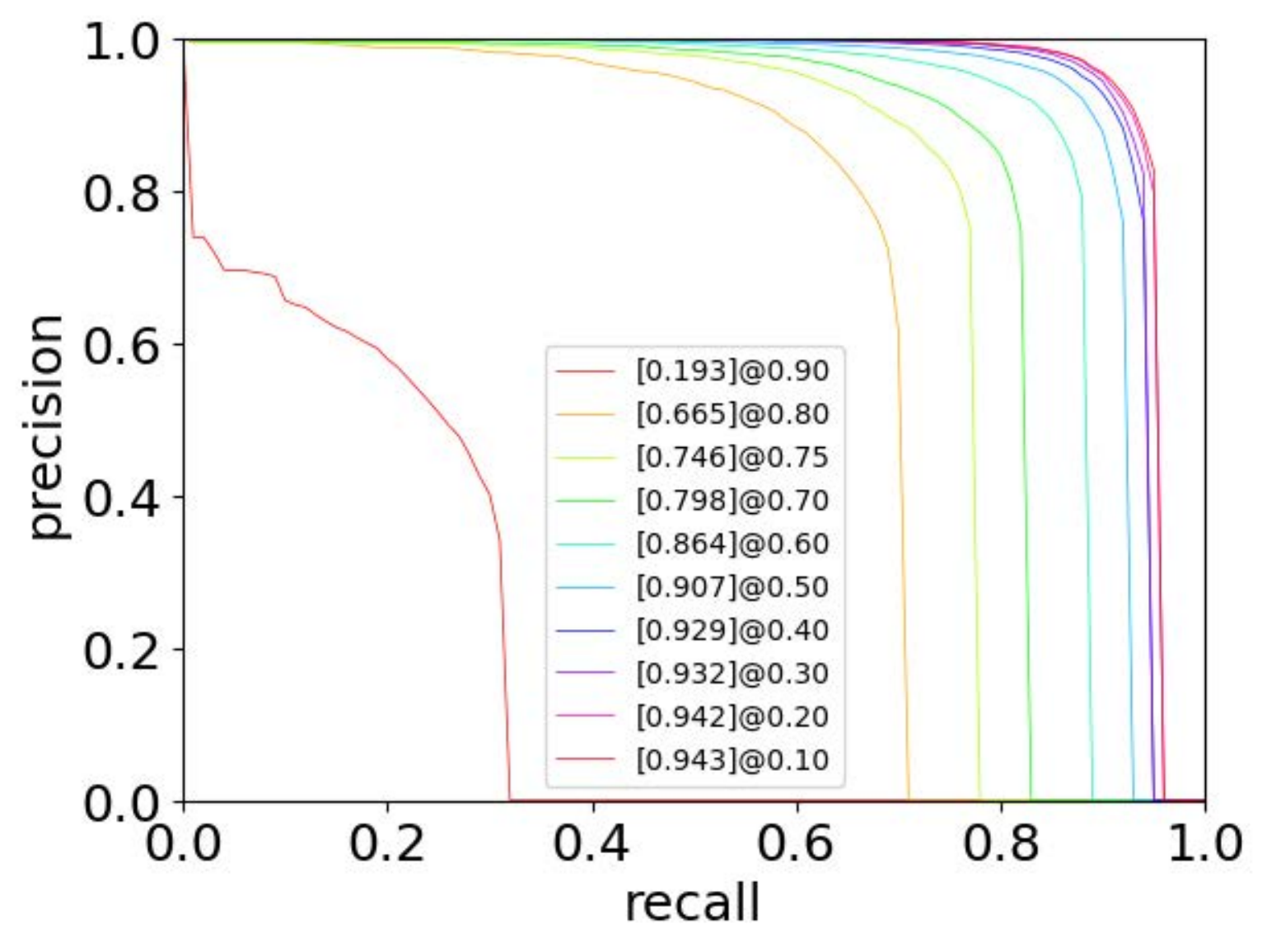}}}\hfill
\subfloat[\centering Random init. ViT ]{{\includegraphics[width=0.3\textwidth]{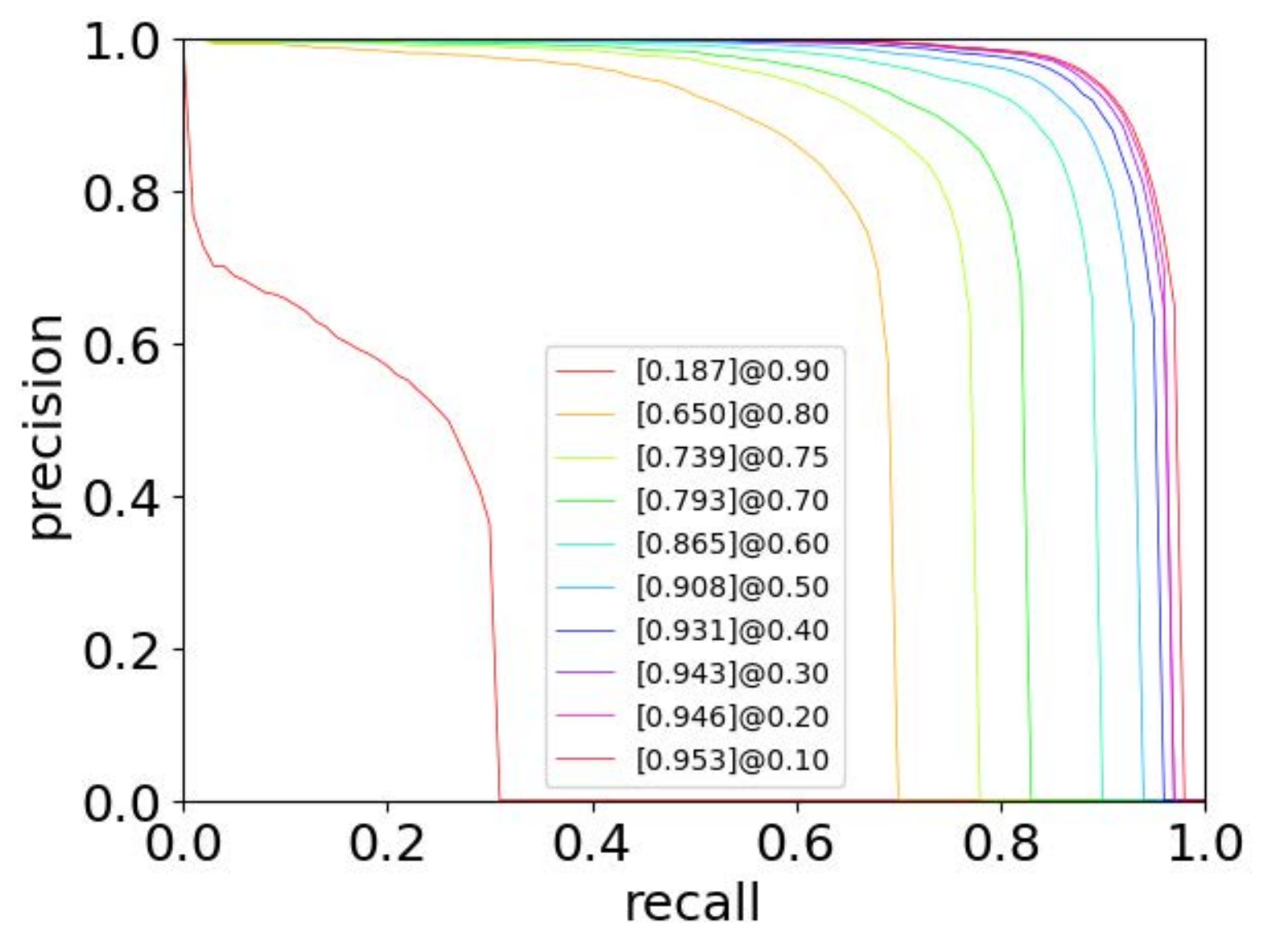}}}%
\caption{Bounding box Precision-Recall curve at different IoU threshold of DAMA and other SSL methods. DAMA has the best scores at the IoU from $0.1:0.75$ and are competitive numbers at $0.8:0.9$.} 
\label{fig:bboxiou}
\end{figure*}
\begin{figure*}[ht]
\centering
\subfloat[\centering DAMA ]{{\includegraphics[width=0.3\textwidth]{figs/iou/segm-allclass-allarea_dama10.pdf}}}\hfill
\subfloat[\centering MAE ]{{\includegraphics[width=0.3\textwidth]{figs/iou/segm-allclass-allarea_mae16.pdf}}}\hfill
\subfloat[\centering MoCo ]{{\includegraphics[width=0.3\textwidth]{figs/iou/segm-allclass-allarea_moco10.pdf}}} \\%
\subfloat[\centering SimMIM-ViT ]{{\includegraphics[width=0.3\textwidth]{figs/iou/segm-allclass-allarea_simmimvit.pdf}}}\hfill
\subfloat[\centering SimMIM-Swin ]{{\includegraphics[width=0.3\textwidth]{figs/iou/segm-allclass-allarea_simmimswin.pdf}}}\hfill
\subfloat[\centering Random init. ViT ]{{\includegraphics[width=0.3\textwidth]{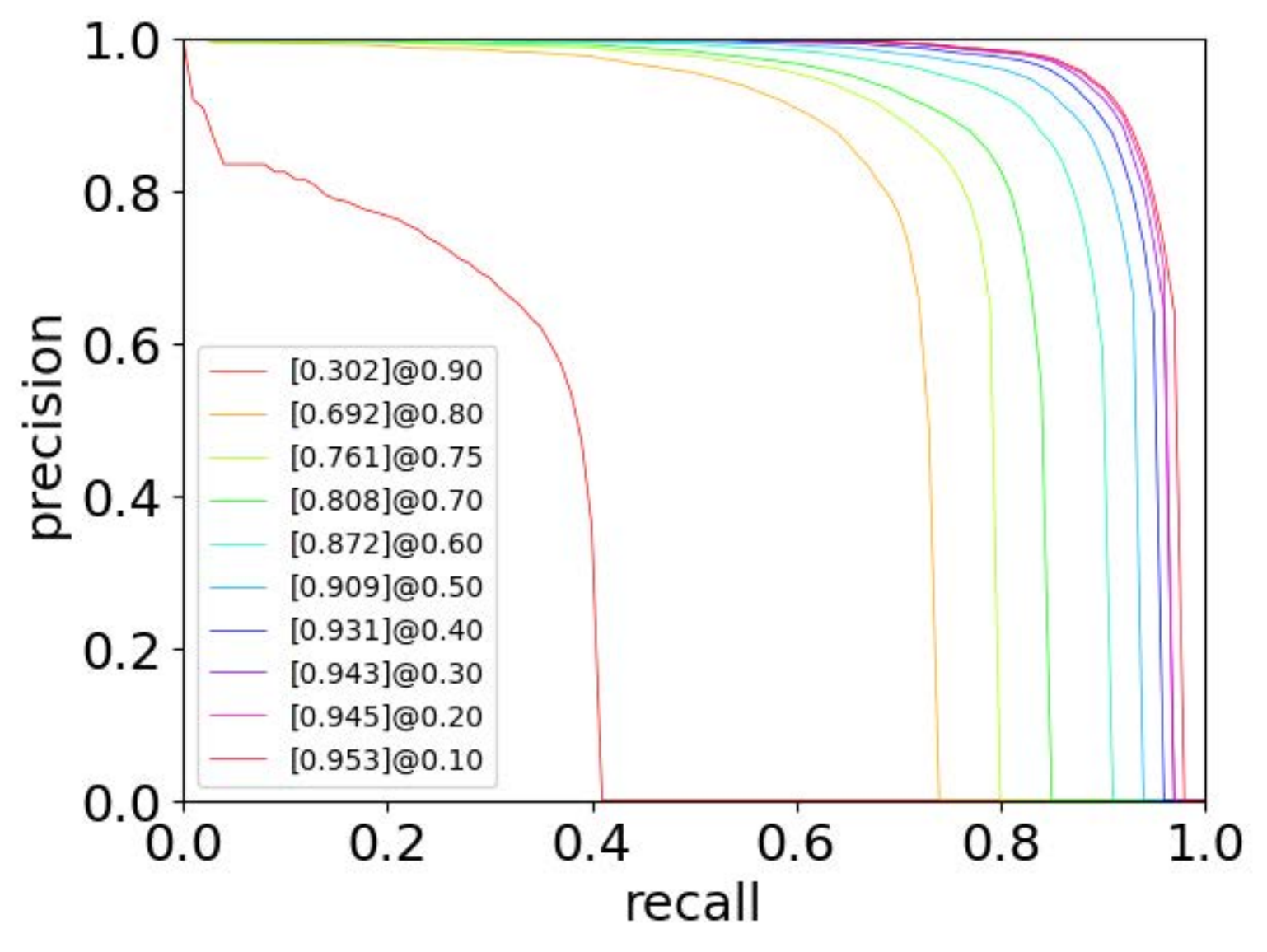}}}%
\caption{Segmentation mask Precision-Recall curve at different IoU threshold of DAMA and other SSL methods. DAMA has the best scores at the IoU from $0.1:0.75$ and are competitive numbers at $0.8:0.9$.} 
\label{fig:segiou}
\end{figure*}
\begin{table*}[t]
\centering
\label{tab:ourparams}
\resizebox{\linewidth}{!}{%
\begin{tabular}{l|cllll|c}
\multicolumn{1}{c|}{Config} & \hspace*{1.75cm}Value\hspace*{1.75cm} &  &  &  & \multicolumn{1}{c|}{Config} & Value \\ 
\cline{1-2}\cline{6-7}
image size & 128$\times$128$\times$7 &  &  &  & image size & 128$\times$128$\times$7 \\
patch size & 16$\times$16 &  &  &  & patch size & 16$\times$16 \\
batch size & 512 &  &  &  & batch size & 512 \\
epochs & 500 &  &  &  & epochs & 150 \\
optimizer & Adam &  &  &  & optimizer & Adam \\
base learning rate & 1.5e-04 &  &  &  & Base learning rate & 1e-02 \\
min learning rate & 0 &  &  &  & min learning rate & 1e-5 \\
weight decay & 0.05 &  &  &  & weight decay & 0.05 \\
learning rate schedule & cosine decay &  &  &  & learning rate schedule & cosine decay \\
warmup epochs & 40 &  &  &  & warmup epochs & 5 \\
augmentation & RandomResizedCrop &  &  &  & augmentation & RandomResizedCrop \\
K-blocks/$\beta$ & 6/2 &  &  &  & droppath/reprob/mixup/cutmix & 0.1/0.25/0.8/1.0 \\
\end{tabular}
}
\caption{Pretraining (left) and finetune (right) setting of our DAMA.}
\end{table*}
\end{document}